\documentclass[10pt,twocolumn,letterpaper]{article}

\usepackage[pagenumbers]{cvpr} 

\usepackage[dvipsnames]{xcolor}         
\usepackage{colortbl}       
\usepackage{booktabs}       
\usepackage{multirow}       
\usepackage{array}          
\usepackage{caption}
\usepackage{algorithm}
\usepackage{algorithmic}
\usepackage{footnote}
\usepackage[accsupp]{axessibility}
\definecolor{cvprblue}{rgb}{0.21,0.49,0.74}
\usepackage[pagebackref,breaklinks,colorlinks,allcolors=cvprblue]{hyperref}

\def\confName{CVPR}
\def\confYear{2026}

\definecolor{lightyellow}{RGB}{255,255,224}
\definecolor{lightpurple}{RGB}{240,230,255}

\definecolor{lightgreen}{RGB}{205,255,205}
\definecolor{lightergreen}{RGB}{235,255,235}
\definecolor{lightred}{RGB}{255,215,215}
\definecolor{lighterred}{RGB}{255,235,235}
\definecolor{lightblue}{RGB}{210,240,255}

\definecolor{darkyellow}{RGB}{153,102,0}

\title{Resolving the Identity Crisis in Text-to-Image Generation}

\author{Shubhankar Borse\thanks{Corresponding Authors. Equal Contribution.} \quad Farzad Farhadzadeh$^*$ \quad Munawar Hayat \quad Fatih Porikli \\
Qualcomm AI Research\thanks{Qualcomm AI Research is an initiative of Qualcomm Technologies, Inc.}\\
  $^*$\texttt{\{sborse, ffarhadz\}@qti.qualcomm.com}\\
  }

\begin{document}

\twocolumn[{%
  \maketitle

  \begin{center}
    \includegraphics[width=0.95\textwidth]{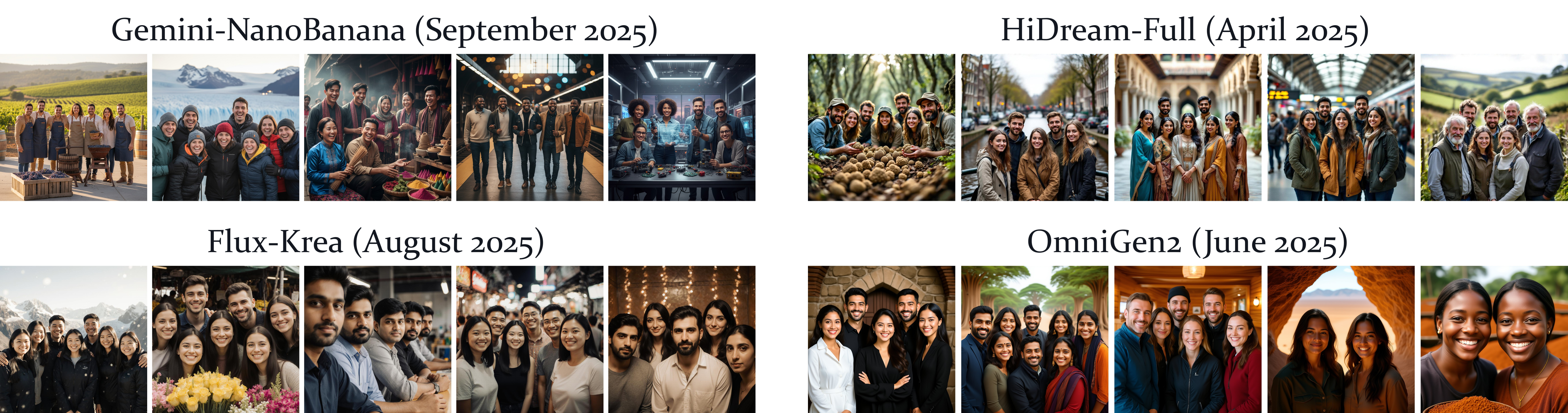}
    \vspace{-2mm}
    \captionof{figure}{{\textbf{The Identity Crisis.} Observe the images carefully. They have been generated by recent SOTA text-to-image methods. From an initial glance, they look great. However, can you spot the issue?}}
    \label{fig:identity_crisis}
  \end{center}
}]

\begingroup
  \renewcommand\thefootnote{\fnsymbol{footnote}}
  \footnotetext[1]{Corresponding authors. Equal contribution.}
  \footnotetext[2]{Qualcomm AI Research, an initiative of Qualcomm Technologies, Inc.}
\endgroup

\begin{abstract}
State-of-the-art text-to-image models suffer from a persistent \textbf{identity crisis} when generating scenes with multiple humans: producing duplicate faces, merging identities, and miscounting individuals. We present \textsc{DisCo} (\textit{Reinforcement with \underline{Di}ver\underline{s}ity \underline{Co}nstraints}), a reinforcement learning framework that directly optimizes identity diversity both within images and \textit{across} groups of generated samples. \textsc{DisCo} fine-tunes flow-matching models using Group-Relative Policy Optimization (GRPO), guided by a compositional reward that: (i) penalizes facial similarity within images, (ii) discourages identity repetition across samples, (iii) enforces accurate person counts, and (iv) preserves visual fidelity and prompt alignment via human preference scores. A single-stage curriculum stabilizes training as prompt complexity increases. Importantly, this method does not require any real data. On the DiverseHumans Testset, \textsc{DisCo} achieves \textbf{98.6\% Unique Face Accuracy} and near-perfect \textbf{Global Identity Spread}, outperforming open-source and proprietary models (e.g., Gemini, GPT-Image) while maintaining perceptual quality. Our results establish cross-sample diversity as a critical axis for resolving identity collapse, positioning \textsc{DisCo} as a scalable, annotation-free solution for multi-human image synthesis. Project page: \url{https://qualcomm-ai-research.github.io/disco/}.
\end{abstract}

\vspace{-2em}
\section{Introduction}
\label{sec:intro}
Text-to-image models have recently achieved impressive realism and controllability, powered by diffusion models~\citep{DiffusionHo, Rombach_2022_CVPR, podell2024sdxl} and flow-based training schemes such as rectified flow and flow matching~\citep{liu2022rectifiedflow, lipman2023flow}. However, when tasked with generating \emph{scenes with multiple people}, current methods frequently replicate nearly identical faces or miscount individuals, undermining realism and limiting practical utility. This limitation was recently pointed out in~\cite{borse2024multihuman}. This is a severe constraint in synthetic data generation for various applications such as training group photo personalization models, consistent character generation and storytelling, narrative media, educational content creation, and simulation environments for social interaction research. As illustrated in Fig.~\ref{fig:identity_crisis}, these failures persist even when overall image quality is high, revealing a bottleneck in \emph{identity differentiation} within and across generations. We term this fundamental issue as the \emph{identity crisis}.

\begin{figure*}[h]
    \centering
    \includegraphics[width=\textwidth]{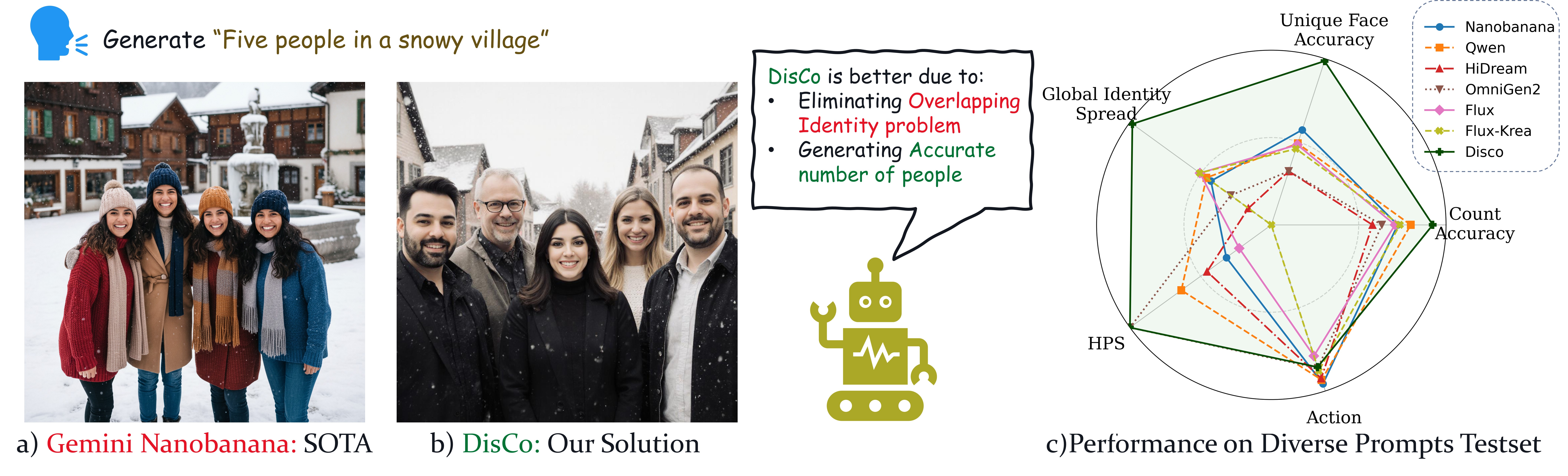}
    \caption{\textbf{\textsc{DisCo} enables better multi-human generation.}
(a) SOTA methods often produce duplicate or inconsistent faces, while (b) \textsc{DisCo} generates distinct, diverse identities. (c) Quantitative results show clear gains in Count Accuracy, Unique Face Accuracy, Identity Spread, and Overall quality(HPSv2 score).}\vspace{-0.5em}
    \label{fig:teaser}
\end{figure*}

Existing text-to-image methods rely mainly on generating realistic and aesthetically pleasing humans~\citep{flux1krea2024, hidreami1technicalreport}. These models do not address identity diversity, especially as the number of people and scene complexity increase. Reinforcement learning (RL) has been applied to the above models to optimize non-differentiable objectives such as prompt adherence, aesthetics, or human preferences~\citep{black2023ddpo, lee2023preference, yang2024d3po}, and GRPO-style algorithms have improved stability and sample efficiency for flow-matching models~\citep{liu2025flowgrpo, danceGRPO2025}. Additionally, RL has shown the ability to correct problematic behaviors that may be ingrained in large models through limited or biased training data. However, \emph{no prior approach explicitly optimizes human-identity diversity both within a single image or across groups of generations for the same prompt}. Beyond the straightforward application of RL, our work makes several novel observations: (1)~applying intra-image diversity rewards alone causes \emph{global diversity collapse}, as duplicate identities shift across samples rather than being eliminated; (2)~diversity optimization introduces reward hacking in the form of face undercounting and grid artifacts, requiring explicit regularization; and (3)~specialist models require structured curriculum learning to converge stably on complex multi-human prompts.

\textbf{We introduce \textsc{DisCo}: Reinforcement with \underline{Di}ver\underline{S}ity \underline{Co}nstraints. It is a novel, sample-efficient RL framework that directly targets identity diversity in multi-human generation, with a particular focus on both intra-image and cross-sample uniqueness.} \textsc{DisCo} explicitly optimizes for diversity not only within a specific image, but also across groups of generated samples, mitigating identity repetition and promoting broader coverage of human facial identities. It fine-tunes flow-matching text-to-image models using Group-Relative Policy Optimization (GRPO)~\citep{liu2025flowgrpo, danceGRPO2025}, guided by a compositional reward that: (i) penalizes facial similarity within images, (ii) discourages identity recurrence across samples, (iii) enforces accurate person counts, and (iv) preserves text-image alignment and image quality via an HPS score~\citep{ma2025hpsv3}, which we find also reinforces the base model's ability to follow fine-grained per-person compositional prompts such as clothing and hairstyle specifications. Reinforcement learning enables flexible optimization over heterogeneous, non-differentiable objectives. It overcomes the rigidity of supervised fine-tuning, which typically requires large annotated datasets. To ensure robustness as prompt complexity increases, \textsc{DisCo} employs a single-stage curriculum that anneals the prompt distribution from simple to diverse multi-human scenarios~\citep{liang2024curriculum}.

\textbf{Empirically, \textsc{DisCo} sets a new standard for multi-human generation by resolving identity collapse both within and across samples.} It significantly reduces identity duplication and improves cross-sample diversity and fidelity across varied prompts and model backbones (e.g., SDXL/SD3.5, FLUX variants, proprietary models), \emph{without requiring auxiliary annotations}. Beyond faces, \textsc{DisCo}'s reward framework generalizes to other visual attributes: training on facial diversity alone improves clothing and body-type variation as a byproduct, and performance on general compositional benchmarks is retained or improved. On DiverseHumans and MultiHuman-TestBench, \textsc{DisCo} consistently improves \emph{Count Accuracy}, \emph{Unique-Faces}, and \emph{Global Identity Spread}, while maintaining perceptual quality.

\begin{itemize}
\item \textbf{Reinforcement for Diversity:} To our knowledge, \textsc{DisCo} is one of the first methods to use reinforcement learning to directly optimize identity diversity in text-to-image generation. We use RL to fine-tune multi-human generation models with identity diversity, quality, prompt-alignment and count-based rewards.
\item \textbf{Group-wise diversity reward:} We uncover a novel finding: \textit{reinforcing intra-image diversity alone does not eliminate duplicate identities, as these duplicates silently spread across samples, causing global identity spread to collapse.} To address this, we introduce a novel group-relative counterfactual reward (Eq.~\ref{eq:group}), directly discouraging cross-sample identity repetitions under GRPO.
\item \textbf{Reward hacking characterizations:} Diversity optimization alone turns out to be gameable, as the model learns to generate fewer people and produce unnatural grid-like face arrangements that technically satisfy the reward while looking nothing like a real scene or following the prompt description. We quantify these failure modes and successfully address them through explicit count-control, perceptual quality and prompt-alignment rewards.
\item \textbf{Single-stage curriculum:} We propose a lightweight sampling curriculum which improves stability and generalization as the number of requested individuals scales, and proves particularly critical for specialist models compared to generalist models.
\item \textbf{State-of-the-art cross-sample identity diversity preserving image quality:} \textsc{DisCo} delivers large gains in identity uniqueness and count accuracy across models and prompts, surpassing proprietary systems including GPT-Image-1 on Unique Face Accuracy (98.6\% vs.\ 85.1\%) and Global Identity Spread (98.3\% vs.\ 89.8\%) \emph{without extra spatial or semantic annotations}.
\end{itemize}

\section{Related Work}
\label{sec:related_work}
\paragraph{Text-to-Image Generation.}
Diffusion models \citep{DiffusionHo} and latent diffusion \citep{Rombach_2022_CVPR,podell2024sdxl} have established high-fidelity text-to-image synthesis. Flow-based formulations, such as rectified flow and flow matching, enable efficient, deterministic sampling with strong quality \citep{liu2022rectifiedflow,lipman2023flow, flux2024, flux1krea2024, hidreami1technicalreport}. Unified multimodal transformers integrate text and image tokens for subject-driven or reference-conditioned generation \citep{xiao2024omnigen,xie2025showo, mao2025ace, OpenAIDocsImageGen, wu2025qwenimagetechnicalreport}. Despite these advances in realism and prompt alignment, \emph{multi-human identity differentiation} remains a persistent failure mode in unconstrained scenes. 

\vspace{-1em}
\paragraph{Multi-Human Generation.} A NeurIPS 2025 study~\cite{borse2024multihuman} discusses the limitations the above methods on the multi-human generation task. They also identify the bias in Human generation by these models, also pointed out by~\cite{chauhan2024identifying}. In their future work section, they observed that current SOTA methods merge identities, repeat faces, or miscount people. These are the precise issues \textsc{DisCo} targets (Fig.~\ref{fig:identity_crisis}).
\vspace{-1em}
\paragraph{Reinforcement Learning for Generative Image Models.}
RL and preference-optimization have been used to optimize non-differentiable objectives such as prompt faithfulness, aesthetics, and human preferences \citep{black2023ddpo, lee2023preference, yang2024d3po}. In the flow-matching setting, GRPO provides value-free, group-relative variance reduction and KL-controlled updates, with curriculum and multi-objective extensions to improve stability and diversity \citep{liu2025flowgrpo,danceGRPO2025}. In contrast to prior work that largely optimizes faithfulness, \textbf{\textsc{DisCo} explicitly encodes identity diversity constraints both intra-image and inter-image}, paired with an identity-aware curriculum, yielding robust gains in multi-human scenes while maintaining quality. A concurrent work~\cite{kazimi2025diverse} also uses RL for diverse video generation, updating the prompt space of video models.

\section{Method}
\label{sec:mathod}
In this Section, we discuss our proposed \textsc{DisCo} finetuning approach in detail. We establish the mathematical foundations in Section~\ref{sec:prelim}. Section~\ref{sec:reward} introduces our proposed compositional reward function. Section~\ref{sec:curriculum} presents a single-stage curriculum learning strategy that gradually transitions from simple to complex multi-person scenarios.

\subsection{Preliminaries}
\label{sec:prelim}
\textbf{Notation.} Let $c$ denote a text prompt, $t \in [0,1]$ index the sampling trajectory from noise ($t{=}1$) to data ($t{=}0$), $\{t_k\}_{k=0}^K$ the discrete time grid with $t_0{=}1 > \cdots > t_K{=}0$, $x_{t_k}$ the latent at step $k$, and $p_t(x)$ the latent distribution at time $t$.

\textbf{Flow-GRPO.} We follow Flow-GRPO~\citep{liu2025flowgrpo} in casting the denoising process as a Markov Decision Process (MDP). Given a text prompt $c$, the state at step $k$ is $s_k = (c, t_k, x_{t_k})$, the action is the next latent $a_k = x_{t_{k+1}}$, and the policy is $\pi_\theta(a_k \mid s_k) = p_\theta(x_{t_{k+1}} \mid x_{t_k}, c)$. A terminal reward $R(s_K) = r(x_{t_K}, c)$ is computed on the final image at $t_K = 0$~\citep{black2023ddpo, yang2024d3po}. To enable exploration while preserving the marginal distributions $\{p_t\}$, the deterministic probability-flow ODE is replaced with a marginal-preserving It\^{o} SDE with controlled stochasticity; we refer the reader to~\citet{liu2025flowgrpo} for full derivations. For each prompt $c$, we sample a group $G = \{\tau_i\}_{i=1}^M$ of $M$ trajectories and compute group-normalized advantages:
\begin{equation}
\vspace{-0.5em}
\tilde{A}_i = \frac{r(\tau_i, c) - \mu_c}{\sigma_c + \epsilon},
\label{eq:adv}
\vspace{-0.5em}
\end{equation}

where $\mu_c$ and $\sigma_c$ are the mean and standard deviation of rewards within the group. The GRPO objective is then:
\begin{equation}
\begin{aligned}
\vspace{-0.5em}
\max_\theta\; &\mathbb{E}_c\!\Bigg[\frac{1}{M}\sum_{i=1}^M
\tilde{A}_i \log \pi_\theta(\tau_i\mid c)\Bigg] \\
&\;-\; \beta_{KL}\,\mathbb{E}_c\!\Big[\mathrm{KL}\big(
\pi_\theta(\cdot\mid c)\,\|\,\pi_{\theta_{\text{ref}}}(\cdot\mid c)\big)\Big],
\end{aligned}
\label{eq:grpo}
\vspace{-0.5em}
\end{equation}
where $\pi_{\theta_{\mathrm{ref}}}$ is the frozen base model and $\beta_{KL}$ controls drift from the reference policy. For efficiency, we train with fewer denoising steps ($K_{\mathrm{train}} \ll K_{\mathrm{test}}$); the full schedule is used at test time.

\begin{figure*}[t]
    \centering
    \includegraphics[width=\textwidth]{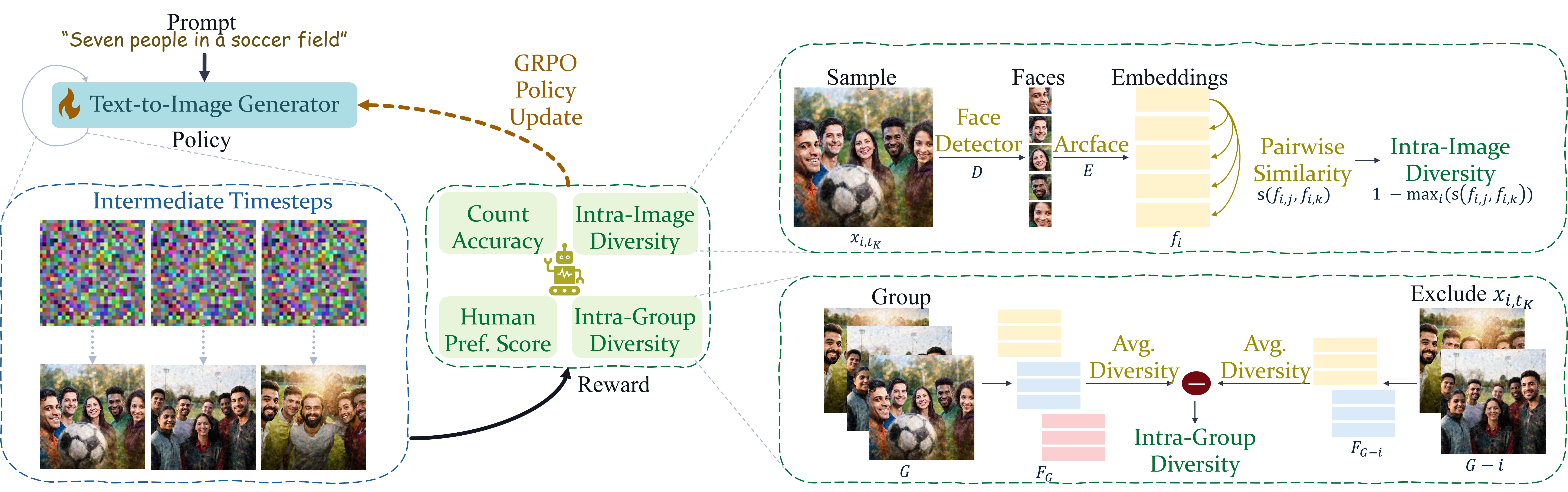} 
    \caption{\textbf{\textsc{DisCo} training overview.} Our method fine-tunes text-to-image models using Flow-GRPO with a compositional reward. Given a prompt, the model generates a group of images evaluated by four components: (1) \textit{Intra-Image Diversity} penalizes duplicate identities within images, (2) \textit{Group-wise Diversity} promotes variation across the group, (3) \textit{Count Accuracy} enforces correct person count, and (4) \textit{HPS Quality} ensures prompt alignment and quality. The combined reward guides GRPO updates to improve identity diversity.}
    \vspace{-1em}
    \label{fig:method}
\end{figure*}

\subsection{Reward Signal}
\label{sec:reward}

Our goal is to train identity-aware generators that (i) avoid duplicate identities within an image, (ii) discourage reusing the same identity across samples of the same prompt, (iii) produce the requested person count, and (iv) preserve text-image quality/alignment. We therefore optimize a compositional reward evaluated at both image- and group-level. Given a prompt $c$ and a group $G=\{\tau_i\}_{i=1}^M$ of trajectories, the terminal image of trajectory $i$ is $x_i\equiv x_{i,t_K}$ and the total reward is
\begin{equation}
\begin{aligned}
r(\tau_i, c, G) =\;& \alpha\,r^{d}_{\text{img}}(x_i)
+ \beta\,r^{d}_{\text{grp}}(x_i, G) \\
&+ \gamma\,r^{c}_{\text{img}}(x_i)
+ \zeta\,r^{q}_{\text{img}}(x_i).
\end{aligned}
\label{eq:totalreward}
\end{equation}
with $\alpha,\beta,\gamma,\zeta>0$. Unless stated otherwise, all four components are bounded in $[0,1]$ to ensure a stable scale under GRPO. We detail each term below, highlighting robustness choices.

\vspace{-1em}
\paragraph{Computing Facial Embeddings.}
Each image \( x_i \) is processed with RetinaFace \cite{deng2019retinafacesinglestagedenseface} Detector $D$, using a confidence threshold \( \eta_{\mathrm{det}} = 0.7 \), yielding bounding boxes \( B_i = \{b_{i,j}\}_{j=1}^{m_i} \). Each face crop, i.e. \( \mathrm{crop}(x_i, b_{i,j}) \) is encoded via ArcFace \cite{Deng_2022} encoder $E$ to produce a \( d \)-dimensional embedding:
\[
f_{i,j} = E\big(\mathrm{crop}(x_i, b_{i,j})\big) \in \mathbb{R}^d.
\]
We denote the set of embeddings for image \( i \) by \( F_i = \{f_{i,1}, \dots, f_{i,m_i}\} \). Identity similarity between embeddings \( u, v \in \mathbb{R}^d \) is computed using cosine similarity \( s(u,v) = \frac{u^\top v}{\|u\|_2 \|v\|_2} \), which simplifies to \( u^\top v \) for \( \ell_2 \)-normalized vectors. Similarity is denoted by \( s(\cdot,\cdot) \) unless otherwise noted.

\vspace{-1em}
\paragraph{Intra-Image Diversity $r_{\mathrm{img}}^{d}$.}
This component utilizes $\{F_i\}$ to enforce diversity by ensuring that the same individual does not appear multiple times within a single generated image.
\begin{equation}
r_{\text{img}}^d(x_i) = \begin{cases}
1 - \max_{j \neq k} s(f_{i,j}, f_{i,k}) & \text{if } m_i \geq 2 \\
0.5 & \text{if } m_i < 2
\end{cases}
\label{eq:intra}
\end{equation}

\vspace{-1.5em}
\paragraph{Group-wise diversity $r_{\mathrm{grp}}^{d}$.} Using this reward, we aim to discourage identity repetition across the group $G$ generated for the same prompt $c$. As the reward needs to be assigned per-image and not per-group, we compute the counterfactual “remove-one” statistic for every image $i$. Let $F_{G}=\bigcup_{i=1}^{M} F_i$ denote all faces across the group. Define
\begin{equation}
\begin{aligned}
S_{G} &= \mathrm{AvgPairwiseSim}(F_{G}) \\
&= \frac{2}{|F_{G}|(|F_{G}|-1)} 
\sum_{\substack{i,j \in \{1,\dots,|F_{G}|\} \\ i<j}} s(f_i, f_j), \; S_{G} \in [0,1].
\end{aligned}
\end{equation}
For image $i$, we remove its faces to get $F_{G-i}$ and compute $S_{G-i}=\mathrm{AvgPairwiseSim}(F_{G-i})$. We define the contribution $\Delta_i = S_{G} - S_{G-i}$. If $S_{G-i} > S_{G}$ then $\Delta_i<0$, meaning $i$ \emph{increases} group diversity; we reward such samples. We map to $[0,1]$ via
\begin{equation}
r_{\text{grp}}^{d}(x_i,G)=\sigma\!\big(-\lambda\,\Delta_i\big),\quad \sigma(u)=\tfrac{1}{1+e^{-u}},\;\; \lambda=5
\label{eq:group}
\end{equation}
Pseudocode is provided in Appendix~\ref{sec:group_algo}. We observe the model performance generally increases when tuned with $r_{\text{img}}^{d}$ and $r_{\text{grp}}^{d}$. However, this model might be susceptible to \textbf{reward hacking}. The nature of hacking, illustrated in Appendix ~\ref{fig:reward_hacking}, includes “grid” artifacts, lesser adherence to prompt and generating lesser number of humans. Hence, we propose methods to regularize against them.

\vspace{-1em}
\paragraph{Count Control $r_{\mathrm{img}}^{c}$.}
To ensure the appropriate number of distinct people and prevent generation of lesser faces, we use face count as a reward:
\begin{equation}
r_{\text{img}}^c(x_i) = \begin{cases}
1 & \text{if } m_i= N_{\text{target}} \\
0 & \text{if } m_i \neq N_{\text{target}}
\end{cases}
\label{eq:count}
\end{equation}
where $N_{\text{target}}$ is number of people in the prompt and $m_i$ is the number of faces detected.

\vspace{-1em}
\paragraph{Quality/Alignment $r_{\mathrm{img}}^{q}$.}
To prevent grid artifacts, lesser prompt adherence \& distorted faces, we use HPSv3~\cite{ma2025hpsv3} as a reward. We normalize \textsc{HPSv3} score to $[0,1]$:
\begin{equation}
r_{\text{img}}^{q}(x_i)=\tilde{q}(x_i)=\frac{\text{HPSv3}(x_i)-q_{\min}}{q_{\max}-q_{\min}},
\label{eq:quality}
\end{equation}
where $q_{\min}=0$ and $q_{\max}=10$. 

Beyond preventing visual artifacts, we find that the HPSv3 reward also preserves and 
improves the base model's compositional instruction-following. Despite no explicit training 
for per-person attribute control, \textsc{DisCo} faithfully follows fine-grained per-person 
specifications such as clothing style and hairstyle alongside identity diversity 
(Appendix~\ref{sec:app_composition}). Paired with results on Geneval2(Appendix~\ref{tab:geneval}), it suggests that optimizing human preference scores
during RL fine-tuning reinforces the base model's general 
prompt-following capability.

\subsection{Single-stage Curriculum Learning}
\label{sec:curriculum}
The difficulty of multi-human generation scales with the number of prompted faces. To handle this complexity, we apply curriculum learning that starts with simple scenarios (2-4 people) and gradually anneals to uniform sampling over the full range (2-$N_{\max}$ people). Let $\{ \mathcal{P}_n \}_{n=2}^{N_{\max}}$ be prompts with $n$ people. Here, $N_{\max}$ is the max number of faces per prompt in training set. The sampling strategy at training step $t$ is:
\begin{equation}
p_t(n) = \begin{cases}
p_{\text{ann}}(n,t) & \text{if } t \leq t_{\text{curriculum}} \\
p_{\text{uni}}(n) & \text{if } t > t_{\text{curriculum}}
\end{cases}
\end{equation}
where the annealing phase interpolates between simple and uniform distributions:
\begin{equation}
\begin{aligned}
p_{\text{ann}}(n,t) &= \lambda_t\,p_{\text{uni}}(n) + (1-\lambda_t)\,p_{\text{simp}}(n),\\
p_{\text{simp}}(n) &= \begin{cases}
\tfrac{1}{3}, & n\in\{2,3,4\}\\[2pt]
0, & \text{otherwise}
\end{cases},\quad
p_{\text{uni}}(n)=\tfrac{1}{N_{\max}-1}.
\end{aligned}
\label{eq:anneal}
\end{equation}
with annealing weight $\lambda_t = \left(\frac{t}{t_{\text{curriculum}}}\right)^{\gamma_c}$, where $\gamma_c > 1$ controls how long the curriculum remains biased toward simple prompts. This strategy ensures a gradual increase in complexity, transitioning from simple to uniform sampling across all prompt types. For further details, refer to Appendix~\ref{sec:curriculum_algo} and~\ref{sec:imp_det}, which describe the curriculum algorithm and hyperparameter settings, respectively.

We apply \textsc{DisCo} finetuning to two models: a \textbf{generalist} (Flux-Dev) model and a \textbf{specialist} (Krea-Dev) model. Generalist models show lesser reliance on curriculum learning due to their broad training on diverse datasets. However, specialist models, optimized for specific aesthetics, benefit significantly from gradual complexity introduction. Curriculum learning is highly effective on the specialist model, as studied in Table~\ref{tab:ablation}.

\subsection{\textsc{DisCo} Algorithm}
\label{sec:overall_algo}
 We provide the complete Pseudocode for DisCo finetuning in the Appendix~\ref{sec:dico_algo}. For each update, we sample $n\sim p_t(\cdot)$, a prompt $c\in \mathcal{P}_n$, generate a group $G$ of $M$ trajectories under the SDE policy, detect faces and compute embeddings, evaluate rewards via Eqs.~\ref{eq:intra}–\ref{eq:quality}, compute advantages via \eqref{eq:adv}, and update $\theta$ with \eqref{eq:grpo}. In the following Section, we discuss the Results of training using DisCo.

\begin{table*}[htbp]    \renewcommand{\arraystretch}{1.5}
    \fontsize{8.5pt}{6.75pt}\selectfont
    \centering
    \begin{tabular}{ll|cccccc}
        \hline
         \multirow{2}{*}{} & \multirow{2}{*}{Model} & \multicolumn{6}{c}{Metrics}\\
        & & Count & Unique Face & Global Identity & HPS & Action & Average \\
        & & Accuracy & Accuracy (UFA) & Spread (GIS) & & Score & \\
        \hline
        \multicolumn{8}{c}{\cellcolor{lightyellow}\textbf{DiverseHumans-TestPrompts} (2-7 People)} \\ 
        
        \hline
        \cellcolor{lightpurple} & \cellcolor{lightpurple} Gemini-Nanobanana & \cellcolor{lightpurple} 72.3 & \cellcolor{lightpurple} 57.2 & \cellcolor{lightpurple} 42.7 & \cellcolor{lightpurple} 31.9 & \cellcolor{lightpurple} 95.7* & \cellcolor{lightpurple} 60.0 \\     
        \multirow{-2}{*}{\cellcolor{lightpurple} Proprietary} & \cellcolor{lightpurple} GPT-Image-1 & \cellcolor{lightpurple} 90.5 &  \cellcolor{lightpurple} 85.1 & \cellcolor{lightpurple} 89.8 & \cellcolor{lightpurple} 33.4 & \cellcolor{lightpurple} 94.5 & \cellcolor{lightpurple} 78.7 \\
        
        \hline
        & HiDream     & \cellcolor{lighterred} 57.9 & 32.3 & \cellcolor{lightred} 16.2 & 32.2 & \cellcolor{lightergreen} 92.4 & \cellcolor{lightred} 46.2 \\
        & Qwen-Image  & 79.8 & 49.0 & 45.9 & 32.6 & \cellcolor{lightgreen} 93.3 & 60.1 \\
        & OmniGen2    & 63.3 & \cellcolor{lighterred} 32.3 & \cellcolor{lighterred} 28.7 & \cellcolor{lightgreen} 33.4 & 86.2 & \cellcolor{lighterred} 48.8 \\
        & DreamO      & 70.5 & \cellcolor{lightred} 31.8 & 35.2 & 32.0 & 82.7 & 50.4 \\
        & SD3.5       & \cellcolor{lightred} 55.3 & 69.1 & 72.5 & \cellcolor{lightred} 28.1 & \cellcolor{lightred} 71.3 & 59.3 \\
        & Flux-Dev        & 70.8 & 48.2 & 50.5 & 31.7 & 78.9 & 56.0 \\
        \multirow{-7}{*}{Open-Source} & Krea-Dev    & 73.6 & 45.8 & 50.6 & \cellcolor{lighterred} 31.2 & 87.9 & 57.8 \\
        
        \hline
        & \textsc{DisCO}(Flux)        & \cellcolor{lightgreen} 92.4 & \cellcolor{lightgreen} 98.6 & \cellcolor{lightgreen} 98.3 & \cellcolor{lightergreen} 33.4 & 85.6 & \cellcolor{lightgreen} 81.7 \\
        \multirow{-2}{*}{Ours} & \textsc{DisCO}(Krea) & \cellcolor{lightergreen} 83.5 & \cellcolor{lightergreen} 89.7 & \cellcolor{lightergreen} 90.6 & 32.2 & 88.2 & \cellcolor{lightergreen} 76.8 \\
        
        \hline
        \multicolumn{8}{c}{\cellcolor{lightyellow}\textbf{MultiHuman-TestBench} (1-5 People)} \\ 
        
        \hline
        \cellcolor{lightpurple} & \cellcolor{lightpurple} Gemini-Nanobanana & \cellcolor{lightpurple} 74.0 & \cellcolor{lightpurple} 67.7 & \cellcolor{lightpurple} 59.7 & \cellcolor{lightpurple} 31.9 & \cellcolor{lightpurple} 98.3* & \cellcolor{lightpurple} 66.3 \\     
        \multirow{-2}{*}{\cellcolor{lightpurple} Proprietary} & \cellcolor{lightpurple} GPT-Image-1 & \cellcolor{lightpurple} 90.7 & \cellcolor{lightpurple} 83.7 & \cellcolor{lightpurple} 81.0 & \cellcolor{lightpurple} 33.2 & \cellcolor{lightpurple} 96.2 & \cellcolor{lightpurple} 77.0 \\
        
        \hline
        & HiDream     & \cellcolor{lightred} 61.1 & \cellcolor{lighterred} 44.8 & \cellcolor{lightred} 22.4 & 32.6 & \cellcolor{lightergreen} 93.6 & \cellcolor{lightred} 50.9 \\
        & Qwen-Image  & \cellcolor{lightergreen} 80.3 & 47.9 & 50.6 & 33.2 & \cellcolor{lightgreen} 94.5 & 61.3 \\
        & OmniGen2    & 74.8 & 45.7 & \cellcolor{lighterred} 36.5 & \cellcolor{lightgreen} 33.5 & 88.2 & \cellcolor{lighterred} 55.7 \\
        & DreamO      & 79.1 & \cellcolor{lightred} 39.0 & 50.4 & 31.8 & 88.6 & 57.8 \\
        & Flux-Dev        & \cellcolor{lighterred} 61.8 & 56.5 & 51.2 & \cellcolor{lighterred} 31.4 & 88.5 & 57.9 \\
        \multirow{-6}{*}{Open-Source} & Krea-Dev    & 67.3 & 52.2 & 55.0 & \cellcolor{lightred} 31.2 & 92.6 & 59.7 \\
        
        \hline
        & \textsc{DisCO}(Flux)        & \cellcolor{lightgreen} 86.6 & \cellcolor{lightgreen} 94.3 & \cellcolor{lightgreen} 88.7 & \cellcolor{lightergreen} 33.3 & 88.9 & \cellcolor{lightgreen} 78.4 \\
        \multirow{-2}{*}{Ours} & \textsc{DisCO}(Krea) & 83.8 & \cellcolor{lightergreen} 80.1 & \cellcolor{lightergreen} 84.1 & 32.9 & 92.3 & \cellcolor{lightergreen} 74.6 \\
           
        \bottomrule
    \end{tabular}
        \caption{Multi-Human Generation Evaluation. Results with * are possibly misleading, as the same MLLM is being probed to perform Generation and act as a judge. \colorbox{lightgreen}{Green} scores indicate the highest results and \colorbox{lightred}{Red} scores indicate the lowest results.}
\label{tab:multihuman_generation}
\end{table*}

\section{Experiment}
\label{sec:results}
\subsection{Experimental Setup}
\subsubsection{Datasets}
\paragraph{Training Data.} For training, we curated a dataset of 30,000 prompts containing group scenes with 2-7 people, with captions generated by GPT-5. The training prompts encompass diverse social contexts, settings, and activities including family gatherings, business meetings, recreational activities, and professional teams to ensure robust multi-human generation capabilities across varied scenarios.

\vspace{-1em}
\paragraph{DiverseHumans.} For evaluation, we developed DiverseHumans, a comprehensive test set of 1,200 prompts systematically organized into six sections of 200 prompts each (corresponding to 2-7 people). Each prompt includes one of four diversity tag variants: no explicit diversity instruction (25\%), general ``diverse faces'' instruction (25\%), single ethnicity specification (25\%), and individual ethnicity assignments for each person (25\%). The dataset deliberately features different contexts from the training set to evaluate generalization capabilities, and for each prompt we generate multiple samples (typically 8-16) to assess both intra-image identity consistency and inter-image diversity.

\vspace{-1em}
\paragraph{MultiHuman-TestBench.} We further evaluate on MultiHuman-TestBench (MHTB), an established recent benchmark introduced at NeurIPS 2025 for multi-human generation. MHTB provides comparison protocols on general multi-human generation capabilities without specific emphasis on identity diversity, and extend the scope of images to people performing simple and complex actions, complementing our DiverseHumans evaluation. Additional details are in the Appendix~\ref{sec:appendix_datasets}.

\subsubsection{Models}
We compare against several baseline models including Nanobanana~\cite{gemini_nanobanana_2024}, SD3.5~\cite{stabilityai2024stable}, FLUX~\cite{flux2024}, Krea~\cite{flux1krea2024}, HiDream-Full~\cite{hidreami1technicalreport}, Qwen-Image~\cite{wu2025qwenimagetechnicalreport}, OmniGen2~\cite{xiao2024omnigen}, DreamO~\cite{mou2025dreamo} and GPT-Image~\cite{OpenAIDocsImageGen}. We fine-tune two open source models, FLUX-Dev(generalist) and Krea-Dev(specialist), using our \textsc{DisCO} framework to allow a direct performance comparison with their baseline counterparts. All implementation details and hyperparameters are provided in the Appendix~\ref{sec:imp_det}.

\subsubsection{Metrics}
To evaluate the performance of our model against the baseline, we report three key metrics: \textbf{Count Accuracy} measures the percentage of generated images that contain the exact number of individuals specified in the prompt. 
\textbf{Unique Face Accuracy (UFA)} quantifies the proportion of images in which all depicted individuals correspond to visually distinct identities, ensuring no duplicates within a single image. 
\textbf{Global Identity Spread (GIS)} is a global metric and assesses identity diversity across a dataset. by computing the ratio of total unique identities to the total prompted identities, in the testset. It indicates how effectively the model avoids repeating the same identities across different images. 
\textbf{HPSv2} assesses image quality and prompt/image alignment. We measure the MLLM \textbf{Action} scores for alignment with textual actions as proposed in MultiHuman-TestBench.
See Appendix~\ref{app:metrics} for the full mathematical details.
\begin{figure*}[t]
    \centering
    \includegraphics[width=\textwidth]{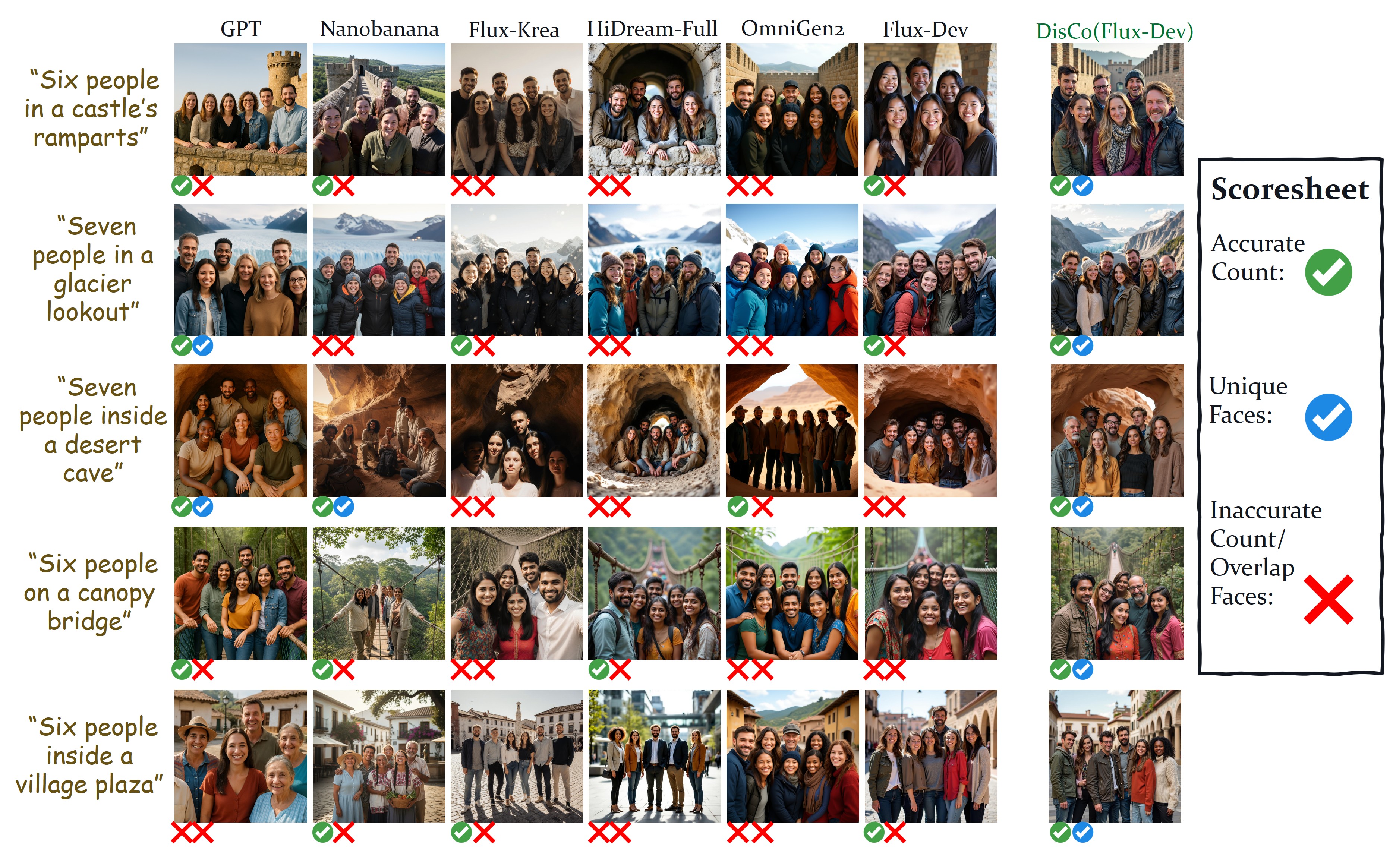}
    \caption{\textbf{\textsc{DisCo} vs. Related Work} \textsc{DisCo} finetuning improves performance over current SOTA methods to consistently generate accurate number of people without overlapping identity. It also maintains high perceptual quality while accurately following input prompts.}
    \label{fig:qualitative_main}
\end{figure*}

\subsection{Results}

\subsubsection{Quantitative Scores}
\paragraph{Diverse Humans Dataset.}
Table~\ref{tab:multihuman_generation} presents comprehensive evaluation results on the DiverseHumans-TestPrompts benchmark. Our \textsc{DisCo} approach demonstrates substantial improvements across all metrics compared to baseline models. \textsc{DisCo}(Flux) achieves 92.4\% Count Accuracy versus baseline Flux's 70.8\%, while \textsc{DisCo}(Krea) reaches 83.5\% compared to Krea's 73.6\%. The most significant gains are in UFA, where \textsc{DisCo}(Flux) reaches 98.6\% versus 48.2\% baseline, and \textsc{DisCo}(Krea) achieves 89.7\% versus 45.8\% baseline. Similarly, Global Identity Spread improves dramatically from 50.5\% to 98.3\% for Flux and from 50.6\% to 90.6\% for Krea. Notably, generalist models like Flux show larger absolute improvements than specialist models like Krea, though both benefit substantially from our approach. Remarkably, \textsc{DisCO}(Flux) surpasses even proprietary models like Nanobanana and GPT-Image-1 in Overall metrics, achieving superior UFA (98.6\% vs 85.1\%) and GIS (98.3\% vs 89.8\%).

\begin{figure*}[t]
  \centering
  \includegraphics[width=\linewidth]{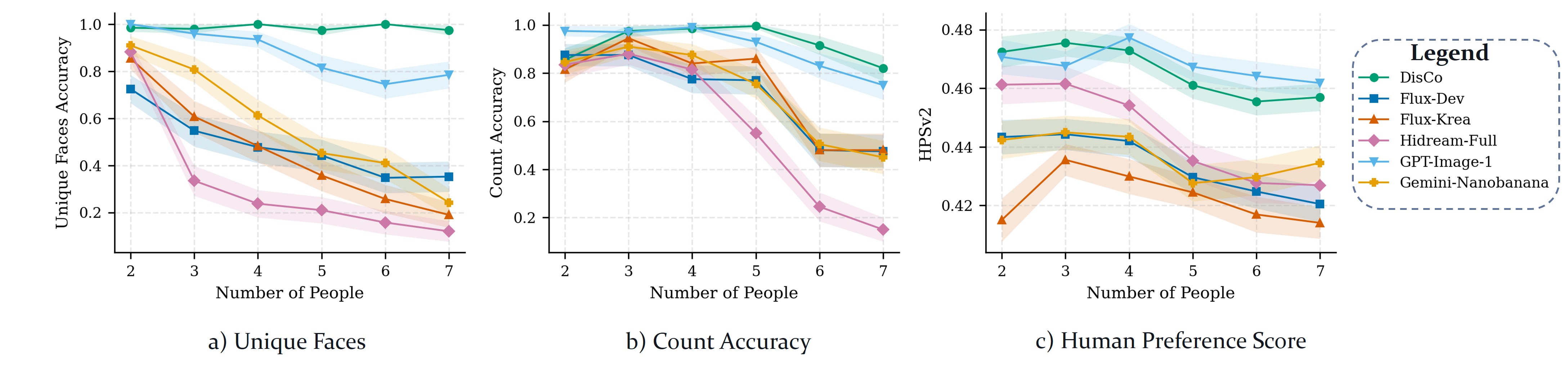}
  \caption{\textbf{Performance vs. number of people.}
We evaluate (a) Unique Face Accuracy, (b) Count Accuracy, and (c) HPSv2 across varying face counts. Error bars show 95\% confidence intervals. \textsc{DisCo}(Flux)in \colorbox{lightgreen}{Green} consistently performs well across all metrics, maintaining high accuracy as face count increases.}
  \label{fig:peoplecount-mosaic}
\end{figure*}

Fig.~\ref{fig:peoplecount-mosaic} illustrates performance across varying numbers of individuals. While baseline models experience significant degradation as complexity increases, \textsc{DisCo} maintains consistently high performance. This robustness is particularly evident in UFA, where \textsc{DisCo} sustains above 90\% accuracy even for scenes with 6-7 individuals, while baseline methods drop below 50\%. This demonstrates \textsc{DisCo}'s superior scalability. In panel (a), UFA performance shows \textsc{DisCo} does not produce overlapping identities even at high person counts, while baseline models exhibit a sharp drop. Panel (b) reveals similar trends for Count Accuracy. Panel (c) confirms that these improvements do not compromise perceptual quality, as HPS scores remain competitive across all configurations.

\vspace{-1em}
\paragraph{MultiHuman-TestBench.}
The MHTB results validate our findings across an independent dataset. \textsc{DisCo}(Flux) achieves 86.6\% Count Accuracy and 94.3\% UFA compared to baseline performance of 61.8\% and 56.5\% respectively, while \textsc{DisCo}(Krea) reaches 83.8\% and 80.1\% versus Krea's 67.3\% and 52.2\%. These consistent improvements across different evaluation protocols demonstrate the generalizability of our approach.

Importantly, over both datasets, HPS quality scores and MLLM Action scores show improvements  over, or remain competitive with the respective (Flux/Krea) baselines. This demonstrates that our identity-focused optimization does not compromise overall generation quality or prompt adherence.

\begin{table*}[htbp]
    \renewcommand{\arraystretch}{1.4}
    \fontsize{8.0pt}{6.75pt}\selectfont
    \centering
    \setlength{\tabcolsep}{4pt}
    \begin{tabular}{l|cccc|c|cccc}
        \hline
        \multirow{2}{*}{Model} & \multicolumn{4}{c|}{Rewards} & \multirow{2}{*}{Curriculum} & \multicolumn{4}{c}{Metrics} \\
        & Image & Group & Count & HPS & & Count & Unique Face & Global Identity & HPS \\
        & Diversity & Diversity & Accuracy & Score & & Accuracy & Accuracy (UFA) & Spread (GIS) & Score \\
        \hline
        
        Krea & & & & & & 73.6 & 45.8 & 50.6 & 31.2 \\
        
        \hline
        
        & \checkmark & & & & & 66.2 & 78.6 & 50.8 & 31.7 \\
        & \checkmark & \checkmark & & & & 67.3 & 80.2 & 72.5 & 32.0 \\
        & \checkmark & \checkmark & \checkmark & & & 81.1 & 83.2 & 68.3 & 31.9 \\
        & \checkmark & \checkmark & \checkmark & \checkmark & & 79.2 & 82.6 & 73.7 & \textbf{32.4} \\
        \rowcolor{lightblue}\multirow{-5}{*}{+DisCo} & \checkmark & \checkmark & \checkmark & \checkmark & \checkmark & \textbf{83.5} & \textbf{89.7} & \textbf{90.6} & 32.2 \\
           
        \bottomrule
    \end{tabular}
    \caption{Ablation Study: Progressive Addition of \textsc{DisCo} Components on Flux-Krea baseline}
    \label{tab:ablation}
\end{table*}

\subsubsection{Qualitative Results}
Fig.~\ref{fig:qualitative_main} showcases the clear visual improvements that \textsc{DisCo} brings to multi-human generation. Where baseline models struggle with repetitive faces and inaccurate person count, our approach delivers different individuals within each scene. Visualizing the examples, several patterns emerge that highlight \textsc{DisCo}'s strengths. Most notably, we see an end to the identity crisis from Fig.~\ref{fig:identity_crisis}, haunting SOTA methods. Instead, \textsc{DisCo} generates individuals with authentic variation in facial features, age, and appearance while preserving the natural demographic diversity we expect in real-world groups. The scenes maintain their coherence and visual appeal.

\subsubsection{Generalizability and Compositional Control}
Beyond facial identity, we demonstrate in Appendix~\ref{app:clothbtypediv} that \textsc{DisCo}'s reward framework generalizes to other visual 
attributes. Replacing the ArcFace reward with a CLIP-based~\cite{radford2021learning} clothing similarity signal 
yields improved clothing diversity across generated individuals. Notably, training 
on facial diversity alone already improves clothing and body-type variation(measured using Qwen3-VL~\cite{bai2025qwen3}) as a 
byproduct. This suggests that the model learns a broader notion of inter-person distinctiveness. On GenEval2~\cite{kamath2025geneval}, \textsc{DisCo} improves over 
baseline Flux across object, counting, positional, and verb understanding attributes, 
demonstrating that diversity-focused RL fine-tuning does not hurt, and in fact 
strengthens general text-to-image capabilities (Appendix~\ref{sec:app_geneval}). 
Finally, despite no explicit training for per-person attribute control, \textsc{DisCo} 
faithfully follows fine-grained compositional prompts specifying clothing style and 
hairstyle per individual, as shown in Appendix~\ref{sec:app_composition}.

\subsection{Ablation Study}
Table~\ref{tab:ablation} ablates individual contributions of each \textsc{DisCo} component. This analysis is conducted on the Krea-Dev baseline, which proved more challenging to converge compared to Flux-Dev. Intra-image diversity dramatically improves unique face accuracy but leaves Global Identity Spread limited, as duplicate identities simply spread across different images rather than being eliminated. Adding group-wise diversity addresses this by enforcing diversity across the entire generation group, substantially improving cross-image identity variation. Count accuracy drops when applying only group-wise rewards due to reward hacking. The model exploits generating fewer people as an easier optimization target. The count control component provides essential regularization, recovering count performance while maintaining identity diversity. However, this introduces perceptual quality issues including unnatural ``grid'' arrangements of faces that technically satisfy requirements but appear artificial. HPS quality control effectively mitigates these artifacts by penalizing obvious visual anomalies. The curriculum learning component delivers substantial improvements. Since Flux-Krea is not a generalist model, training convergence proved challenging without proper task decomposition. Curriculum learning addresses this by progressing from simple to complex scenarios, enabling the specialized model to learn the difficult multi-human generation task incrementally. As evident from the scores, each component contributes meaningfully to the final performance, with the complete framework achieving optimal results across all metrics despite the challenging baseline characteristics. Additional ablation studies, such as varying the curriculum bounds, grid search on reward weights and varying the intra-image aggregation function can be found in Appendix~\ref{sec:ext_res}. We also provide latency trade-offs as well as a user preference study for our proposed metrics.

\vspace{-0.5em}
\section{Conclusion}
\label{sec:conclusions}
\vspace{-0.5em}

In this paper, we uncover and resolve the \textit{identity crisis} of text-to-image models: when asked to generate a crowd, they consistently reuse the same face. We propose \textsc{DisCo}, fine-tuning flow-matching models via GRPO with a compositional reward built around a novel finding: optimizing diversity within an image is not enough, as duplicates simply migrate across samples instead. Our group-relative reward catches this cross-sample collapse, complemented by count accuracy, perceptual quality, and text alignment terms to keep generations grounded. Combined with a single-stage curriculum, \textsc{DisCo} turns a model that copies faces into one that generates distinct individuals at scale, surpassing proprietary models on identity diversity without sacrificing quality or requiring extra annotations. Beyond faces, our rewards generalize to clothing and body-type diversity, and we demonstrate that RL fine-tuning retains general compositional capability. Future directions include extending identity-aware RL to personalized generation, video generation, and richer attribute diversity. Some of these directions are already being explored in follow-up work~\citep{borse2025ar2can} leveraging \textsc{DisCo}.




{
    \small
    \bibliographystyle{ieeenat_fullname}
    \bibliography{cvpr2026/main}
}

\clearpage
\newpage
\onecolumn
\appendix

\counterwithin{figure}{section}
\counterwithin{table}{section}

\renewcommand{\contentsname}{\Large\textbf{Appendices}}
\setcounter{tocdepth}{3}


\tableofcontents



\section{Extended Method}
\label{app:ext_method}
The following algorithms provide detailed pseudocode implementations of the key components described in Section~\ref{sec:mathod}. Algorithm~\ref{alg:group_diversity} formalizes the group-wise diversity computation from Section~\ref{sec:reward}, Algorithm~\ref{alg:curriculum} details the curriculum learning strategy from Section~\ref{sec:curriculum}, and Algorithm~\ref{alg:disco} presents the complete training procedure that integrates all components from Section~\ref{sec:overall_algo}.

\subsection{Group-wise diversity Algorithm}
\label{sec:group_algo}
\begin{algorithm}
\caption{Group-Level Identity Diversity Computation}
\begin{algorithmic}
\REQUIRE Group $G = \{x_i\}_{i=1}^M$, face embeddings $\{F_i\}_{i=1}^M$, scaling parameter $\lambda$
\STATE $F_{G} \leftarrow \bigcup_{i=1}^M F_i$ \COMMENT{All faces across group}
\STATE $S_{G} \leftarrow \text{AvgPairwiseSim}(F_{G})$ \COMMENT{Baseline group similarity}
\FOR{$i = 1$ to $M$}
    \STATE $F_{G-i} \leftarrow F_{G} \setminus F_i$ \COMMENT{Remove faces from image $i$}
    \STATE $S_{G-i} \leftarrow \text{AvgPairwiseSim}(F_{G-i})$ \COMMENT{Similarity without image $i$}
    \STATE $\Delta_i \leftarrow S_{G} - S_{G-i}$ \COMMENT{Image $i$'s contribution to similarity}
    \STATE $r_{\text{grp}}^d(x_i, G) \leftarrow \sigma(-\lambda \cdot \Delta_i)$ \COMMENT{Sigmoid mapping with $\sigma(u) = \frac{1}{1+e^{-u}}$}
\ENDFOR
\RETURN $\{r_{\text{grp}}^d(x_1, G), \ldots, r_{\text{grp}}^d(x_M, G)\}$
\end{algorithmic}
\label{alg:group_diversity}
\end{algorithm}

Algorithm~\ref{alg:group_diversity} provides the implementation details for the counterfactual reward computation described in Section~\ref{sec:reward}. The algorithm efficiently computes the baseline similarity $S_G$ once per group, then performs $M$ leave-one-out evaluations to determine each image's diversity contribution $\Delta_i$. In practice, with typical group sizes of $M = 21$ and face counts of 2-7 per image, the algorithm executes efficiently within the GRPO training loop. 

\subsection{Single-Stage Curriculum Learning Algorithm}
\label{sec:curriculum_algo}
\begin{algorithm}
\caption{\textsc{DisCo}: Single-stage Curriculum Learning}
\begin{algorithmic}
\REQUIRE Prompt sets $\{\mathcal{P}_n\}_{n=2}^{N_{\max}}$, curriculum parameters $t_{\text{curriculum}}$, $\gamma_c$
\STATE Initialize training step $t = 0$
\WHILE{training not converged}
    \IF{$t \leq t_{\text{curriculum}}$}
        \STATE $\lambda_t \leftarrow \left(\frac{t}{t_{\text{curriculum}}}\right)^{\gamma_c}$ \COMMENT{Exponential annealing weight}
        \FOR{$n = 2$ to $N_{\max}$}
            \IF{$n \in \{2,3,4\}$}
                \STATE $p_{\text{simple}}(n) \leftarrow \frac{1}{3}$
            \ELSE
                \STATE $p_{\text{simple}}(n) \leftarrow 0$
            \ENDIF
            \STATE $p_{\text{uniform}}(n) \leftarrow \frac{1}{N_{\max}-1}$
            \STATE $p_t(n) \leftarrow \lambda_t \cdot p_{\text{uniform}}(n) + (1-\lambda_t) \cdot p_{\text{simple}}(n)$
        \ENDFOR
    \ELSE
        \FOR{$n = 2$ to $N_{\max}$}
            \STATE $p_t(n) \leftarrow \frac{1}{N_{\max}-1}$ \COMMENT{Uniform sampling}
        \ENDFOR
    \ENDIF
    \STATE Sample $n \sim p_t(\cdot)$
    \STATE Sample prompt $c$ from $\mathcal{P}_n$
    \STATE Generate group $G$ and update model with prompt $c$
    \STATE $t \leftarrow t + 1$
\ENDWHILE
\end{algorithmic}
\label{alg:curriculum}
\end{algorithm}

Algorithm~\ref{alg:curriculum} provides the implementation details for the exponential curriculum strategy outlined in Section~\ref{sec:curriculum}. The gamma parameter $\gamma_c$ controls the steepness of complexity introduction, with higher values maintaining focus on simple prompts for longer durations before transitioning to the full complexity range. The curriculum duration $t_{\text{curriculum}}$ determines the absolute training steps allocated to gradual complexity introduction before switching to uniform sampling across all prompt types. We define scenarios with 2-4 people as "simple" based on empirical analysis of baseline model performance degradation patterns. As shown in Figure~\ref{fig:peoplecount-mosaic}, both Count Accuracy and Unique Face Accuracy exhibit the most pronounced performance drops at the 4-person threshold, with steeper degradation beyond this point, motivating our curriculum design that focuses initial training on these manageable scenarios before introducing the full complexity range.

\subsection{DisCo Algorithm}
\label{sec:dico_algo}
\begin{algorithm}
\caption{\textsc{DisCo}: Overall Algorithm}
\begin{algorithmic}
\REQUIRE Pretrained flow-matching model $v_{\theta_0}$, prompt dataset $\mathcal{P}$, curriculum parameters $\eta$, $t_{\text{start}}$, $t_{\text{end}}$, reward weights $\alpha, \beta, \gamma, \zeta$

\WHILE{not converged}
    \STATE Sample $n \sim p_t(\cdot)$ and prompt $c \in \mathcal{P}_n$ using Algorithm~\ref{alg:curriculum}
    \STATE Generate group $G = \{\tau_i\}_{i=1}^M$ using SDE policy $\pi_\theta(\cdot | c)$
    \STATE Extract facial embeddings: $F_i = \{E(\text{crop}(x_{i}, b)) : b \in D(x_{i})\}$ for all $i$
    \STATE Compute compositional rewards: $r(\tau_i,G) = \alpha r_{\text{img}}^d + \beta r_{\text{grp}}^d + \gamma r_{\text{img}}^c + \zeta r_{\text{img}}^q$
    \STATE Compute group-normalized advantages $\{\tilde{A}_i\}$ and update $\theta$ using GRPO objective
    \STATE $t \leftarrow t + 1$
\ENDWHILE

\RETURN Fine-tuned model $\theta$
\end{algorithmic}
\label{alg:disco}
\end{algorithm}

Algorithm~\ref{alg:disco} integrates all components described in Section~\ref{sec:mathod} into the complete \textsc{DisCO} training procedure. The reward weights $\alpha, \beta, \gamma, \zeta$ control the relative importance of intra-image diversity, group diversity, count accuracy, and quality objectives respectively, allowing fine-grained control over the optimization priorities. The group size $M$ determines the number of trajectories generated per prompt, directly affecting both the quality of group-normalized advantage estimation and the computational cost per training iteration.

\section{Dataset Details}
\label{sec:appendix_datasets}

\subsection{Training Dataset}
Our training dataset consists of 30,000 carefully curated prompts designed to capture diverse multi-human scenarios. Each prompt describes group scenes containing 2-7 people engaged in various activities and contexts. The captions were generated using GPT-5 to ensure high-quality, diverse descriptions that encompass a wide range of:

\begin{itemize}
    \item \textbf{Social contexts}: Family gatherings, business meetings, friend groups, professional teams, recreational activities
    \item \textbf{Settings}: Indoor and outdoor environments, formal and informal occasions, workplace and leisure contexts
    \item \textbf{Activities}: Collaborative tasks, social interactions, professional activities, recreational pursuits
    \item \textbf{Group compositions}: Varying numbers of individuals (2-7) with diverse demographic representations
\end{itemize}

The prompts were designed to avoid overlap with evaluation datasets while maintaining sufficient diversity to train robust multi-human generation capabilities. The following are 5 examples of these prompts.

\begin{itemize}
  \item {\fontfamily{cmss}\selectfont \textcolor{darkyellow}{Seven people on the desert dunes, hazy sun, diverse faces, clear faces visible, studio-quality, vivid detail}}
  \item {\fontfamily{cmss}\selectfont \textcolor{darkyellow}{Six people in an astronomy studio, Clean composition, Professional portrait, Portrait photography, Soft shadows, Natural lighting, Even exposure}}
  \item {\fontfamily{cmss}\selectfont \textcolor{darkyellow}{Three people in an aviation observatory, Sharp focus, Clean composition, Bokeh background, Color graded, Smiling expressions, Well lit}}
  \item {\fontfamily{cmss}\selectfont \textcolor{darkyellow}{Five people in a dawn-lit bakeshop, Studio quality, Even exposure, Group harmony, Cinematic lighting, Portrait photography, Soft shadows}}
  \item {\fontfamily{cmss}\selectfont \textcolor{darkyellow}{Seven people on a coastal boardwalk, afternoon light, diverse faces, clear faces visible, ultra-realistic, 8K resolution}}
\end{itemize}
\subsection{Evaluation Datasets}

\subsubsection{DiverseHumans Test Set}
We developed DiverseHumans, a comprehensive evaluation dataset of 1,200 prompts specifically designed to assess identity consistency and diversity in multi-human generation. The dataset is systematically organized as follows:

\textbf{Structure}: Six sections of 200 prompts each, corresponding to scenes with 2, 3, 4, 5, 6, and 7 people respectively.

\textbf{Diversity Tags}: Each prompt includes one of four diversity specification levels:
\begin{enumerate}
    \item \textbf{No tag} (25\% of prompts): Basic scene descriptions without explicit diversity instructions
    \item \textbf{``Diverse faces'' tag} (25\% of prompts): General diversity encouragement
    \item \textbf{Single ethnicity specification} (25\% of prompts): Mentions one of six racial/ethnic categories
    \item \textbf{Individual ethnicity assignments} (25\% of prompts): Specific ethnicity assigned to each person
\end{enumerate}

\textbf{Example Prompts}:
\begin{itemize}
    \item \textit{No tag}: {\fontfamily{cmss}\selectfont \textcolor{darkyellow}{Five people on a island cove beach, High dynamic range, Group harmony, Professional portrait, Natural lighting, Smiling expressions}}
    \item \textit{Diverse faces}: {\fontfamily{cmss}\selectfont \textcolor{darkyellow}{Five people in a antique arcade, High dynamic range, Sharp focus, Group harmony, Clear faces, Smiling expressions, Diverse faces among people}}
    \item \textit{Single ethnicity}: {\fontfamily{cmss}\selectfont \textcolor{darkyellow}{Five people in a sidewalk cafe, Sharp focus, Bokeh background, Well lit, Clear faces, Group harmony, Indian ethnicity}}
    \item \textit{Individual assignments}: {\fontfamily{cmss}\selectfont \textcolor{darkyellow}{Five people in a coastal market, Bokeh background, High dynamic range, Sharp focus, Professional portrait, Portrait photography, One person is White, One person is Middle-eastern, One person is Asian, One person is Black, One person is Hispanic}}
\end{itemize}

\textbf{Context Differentiation}: The DiverseHumans prompts deliberately feature different contexts and scenarios compared to the training set to evaluate generalization capabilities and prevent overfitting to training distributions.

\subsubsection{MultiHuman-TestBench (MHTB)}
We additionally evaluate on the established MultiHuman-TestBench, a standardized benchmark for multi-human generation that provides consistent evaluation protocols and enables fair comparison with existing methods. MHTB focuses on general multi-human generation capabilities without specific emphasis on identity diversity, complementing our DiverseHumans evaluation. MHTB also asks for people performing specific actions (cooking, boxing, dancing, etc.) ranging from simple to complex, which is a key differentiator to DiverseHumans testset. We use their official implementation\footnote{\url{https://github.com/Qualcomm-AI-research/MultiHuman-Testbench}} to download data and compute metrics.

\section{Evaluation Metrics}
\label{app:metrics}
To comprehensively evaluate multi-human generation performance as described in Section~\ref{sec:results}, we employ three core metrics that capture different aspects of identity consistency and counting accuracy. All metrics are computed using facial embeddings extracted via RetinaFace detection followed by ArcFace encoding, as detailed in our reward computation pipeline. All metrics are reported as percentages.

\paragraph{Count Accuracy.}
This metric measures the percentage of generated images that contain the exact number of individuals specified in the input prompt. For a given prompt $c$ with target count $N_{\text{target}}(c)$ and evaluation set $\mathcal{X}$, Count Accuracy is defined as:
$$\text{Count Accuracy (\%)} = 100 \times \frac{1}{|\mathcal{X}|} \sum_{x \in \mathcal{X}} \mathbf{1}\{F(x) = N_{\text{target}}(c)\}$$
where $F(x) = |D(x)|$ represents the number of detected faces in image $x$ using RetinaFace with confidence threshold $\kappa_{\text{det}} = 0.7$.

\paragraph{Unique Face Accuracy (UFA).}
This metric quantifies the percentage of images in which all depicted individuals correspond to visually distinct identities, ensuring no duplicate faces within a single image. We define faces as duplicates if their cosine similarity exceeds a threshold. Specifically, within image $x$, duplicates exist if:
$$\exists \, i \neq j : s(f_i, f_j) \geq \kappa_{\text{dup}}$$
where $s(\cdot, \cdot)$ denotes cosine similarity between face embeddings. The UFA metric is then computed as:
$$\text{UFA (\%)} = 100 \times \frac{1}{|\mathcal{X}|} \sum_{x \in \mathcal{X}} \mathbf{1}\{\text{no duplicates in } x\}$$
We set $\kappa_{\text{dup}} = 0.5$.

\paragraph{Global Identity Spread (GIS).}
This metric assesses identity diversity across an entire dataset of generated images by measuring the percentage of unique identities created relative to the total number of people requested across all prompts. 
For a batch $\mathcal{X}$ of images generated from prompts with respective target counts $\{N_{\text{target}}(c_i)\}$, we first cluster all face embeddings $\bigcup_{x \in \mathcal{X}} F(x)$ using single-linkage clustering with threshold $\kappa_{\text{dup}} = 0.5$. Let $C$ denote the total number of unique clusters (identities) found. The Global Identity Spread is then computed as:
$$\text{GIS (\%)} = 100 \times \frac{C}{\sum_{i} N_{\text{target}}(c_i)}$$
where the denominator represents the total number of people requested across all prompts in the batch. Higher GIS values indicate better identity diversity, with perfect diversity yielding GIS = 100\% when every requested person has a unique identity.

\paragraph{Action Score.} We use the Action score as implemented in the MultiHuman-TestBench~\cite{borse2024multihuman} paper. This is an MLLM metric, which prompts Gemini-2.0-Flash using the image, and asks if the people in the image are performing the Action requested by the prompt.

\paragraph{HPSv2:} Due to our use of HPSv3 as a reward, we use the HPSv2 model to measure perceptual quality and prompt alignment. This step is to make the comparison with other methods fair, which may or may not have been trained with an HPSv3 reward.

\section{Implementation Details}
\label{sec:imp_det}

\paragraph{\textsc{DisCo} Training.} We implement \textsc{DisCo} using the public \texttt{flow\_grpo}\footnote{\url{https://github.com/yifan123/flow_grpo}} framework with Flux pipeline, training in bf16 mixed precision on 512×512 images. Training uses 14 timesteps for reward computation and 28 steps for evaluation, with classifier-free guidance of 4.5 for Flux-Krea and 3.5 for Flux-Dev. We train for 480 epochs with batch sizes of 3 (train) and 16 (test), with a group size of 21. The compositional reward function combines intra-image diversity ($\alpha = 0.50$), group-wise diversity ($\beta = 0.10$), count accuracy ($\gamma = 0.15$), and HPS quality ($\zeta = 0.15$) components, with KL regularization weight $\beta_{KL} = 0.01$ to stabilize learning. We apply the proposed curriculum with $t_{\text{curriculum}} = 60$ epochs, and $\gamma_c = 3$

Training is distributed across 21 GPUs on 3 H100 clusters, with a single dedicated GPU for HPSv3 reward (3 nodes, 7 GPUs per node for training, 1 GPU as the HPSv3 server). We use a learning rate of $1 \times 10^{-4}$ with EMA enabled and checkpoint every 30 epochs. The curriculum learning strategy transitions from simple to complex prompts using exponential weighting parameter $\eta = 2.0$, with transition period from steps 10,000 to 40,000. Face detection uses RetinaFace \citep{deng2019retinafacesinglestagedenseface} with confidence threshold 0.7, followed by ArcFace \citep{Deng_2022} embeddings for identity similarity computation. Total training time to 480 epochs is \textbf{13 hours}.

\paragraph{Baseline Model Evaluation Settings.} For fair comparison, we evaluate all baseline models using their recommended hyperparameters from official documentation. For OmniGen2, we use 50 inference steps with text guidance scale of 2.5 and image guidance scale of 3.0 for multi-modal tasks.\footnote{\url{https://huggingface.co/OmniGen2/OmniGen2}} We set FLUX-Dev to 50 timesteps with CFG guidance of 3.5, while for FLUX-Krea we use 28 timesteps with CFG guidance 4.5 as specified in the official repository.\footnote{\url{https://huggingface.co/black-forest-labs/FLUX.1-dev}}\footnote{\url{https://github.com/krea-ai/flux-krea}} For SD3.5-Large, we apply 40 timesteps with guidance scale of 4.5.\footnote{\url{https://huggingface.co/stabilityai/stable-diffusion-3.5-large}} We configure HiDream-I1 Full model with 50 timesteps and guidance scale 5.0.\footnote{\url{https://huggingface.co/HiDream-ai/HiDream-I1-Full}} We use 12 timesteps for DreamO and CFG guidance 4.5.\footnote{\url{https://github.com/bytedance/DreamO}}. We generate all images at 1024×1024 resolution. We set a different seed for every image (the image index itself), and we share these seeds across all evaluations.

\section{Extended Results}
\label{sec:ext_res}

\subsection{Quantitative Results}

The quantitative results presented in this section provide detailed analysis of \textsc{DisCo}'s performance across various experimental conditions and model configurations. These results complement the main paper findings by examining performance variations across different prompt types, reward weight configurations, and computational efficiency metrics.

\subsubsection{Performance on various Diversity Tags in prompts}
Table~\ref{tab:diversity_splits} analyzes performance across the four diversity specification levels in our DiverseHumans dataset. The results reveal interesting patterns that demonstrate \textsc{DisCo}'s effectiveness in addressing different types of diversity challenges. 

For Unique Face Accuracy, baseline models show variable performance across diversity tags, with some models (like Gemini-Nanobanana) performing significantly better on explicit diversity prompts (D=2: 70.8\%, D=4: 78.3\%) compared to unspecified prompts (D=1: 41.5\%). This suggests that baseline models can leverage explicit diversity instructions but struggle with implicit diversity requirements. In contrast, \textsc{DisCo} maintains consistently high UFA performance (97.7-99.7\%) across all diversity specifications, effectively eliminating duplicate identities regardless of prompt formulation.

The Global Identity Spread metric reveals a complementary pattern: baseline models generally achieve higher GIS scores on simpler diversity specifications (D=1, D=3) but struggle with complex individual assignments (D=4), where detailed ethnicity specifications appear to constrain their generation diversity. For instance, Flux-Krea drops from 71.9\% (D=1) to 52.8\% (D=4), and OmniGen2 falls from 48.5\% to 29.2\%. This indicates that explicit individual constraints paradoxically reduce overall identity diversity in baseline models. \textsc{DisCo} overcomes this limitation, achieving near-perfect GIS scores (98.5-100\%) across all prompt types, demonstrating that our compositional reward system successfully handles both implicit and explicit diversity requirements without compromising identity uniqueness.

These patterns confirm that \textsc{DisCo} generalizes robustly across diverse prompt formulations, resolving the fundamental tension between following specific diversity instructions and maintaining overall identity spread that challenges existing models.

\begin{table*}[htbp]
    \caption{Performance across diversity tags (D=1: No tag, D=2: "Diverse faces", D=3: Single ethnicity, D=4: Individual assignments). DisCo shows consistent improvements across all diversity specifications. \colorbox{lightgreen}{Green} scores indicate the highest results and \colorbox{lightred}{Red} scores indicate the lowest results.}
    \renewcommand{\arraystretch}{1.5}
    \fontsize{8.0pt}{6.75pt}\selectfont
    \centering
    \setlength{\tabcolsep}{5pt}
    \begin{tabular}{l|cccc|cccc|cccc}
        \hline
         \multirow{2}{*}{Model} & \multicolumn{4}{c|}{Count Accuracy} & \multicolumn{4}{c|}{Unique Face Accuracy} & \multicolumn{4}{c}{Global Identity Spread} \\
        & D=1 & D=2 & D=3 & D=4 & D=1 & D=2 & D=3 & D=4 & D=1 & D=2 & D=3 & D=4 \\
        \hline
        \multicolumn{13}{c}{\cellcolor{lightyellow}\textbf{DiverseHumans-TestPrompts}} \\ 
        \hline
        Gemini-Nanobanana & 71.0 & \cellcolor{lightergreen} 71.7 & 70.7 & 76.0 & 41.5 & \cellcolor{lightergreen} 70.8 & 38.3 & \cellcolor{lightergreen} 78.3 & 56.6 & \cellcolor{lightergreen} 69.2 & 53.8 & 55.7 \\
        Flux-Dev & 70.0 & 70.0 & 69.0 & 74.3 & 47.8 & 41.7 & \cellcolor{lightergreen} 47.0 & 56.3 & 64.74 & 58.8 & \cellcolor{lightergreen} 67.8 & \cellcolor{lightergreen} 62.3 \\
        Flux-Krea & \cellcolor{lightergreen} 75.0 & 68.0 & \cellcolor{lightergreen} 71.3 & \cellcolor{lightergreen} 80.3 & \cellcolor{lightergreen} 51.3 & 45.3 & 37.5 & 49.2 & \cellcolor{lightergreen} 71.9 & 66.66 & 56.9 & 52.8 \\
        OmniGen2 & \cellcolor{lighterred} 62.3 & \cellcolor{lighterred} 61.3 & \cellcolor{lighterred} 67.0 & \cellcolor{lighterred} 62.3 & \cellcolor{lighterred} 32.3 & 33.5 & \cellcolor{lighterred} 27.2 & \cellcolor{lighterred}  36.2 & \cellcolor{lighterred} 48.5 & \cellcolor{lighterred} 36.2 & \cellcolor{lighterred} 41.2 & \cellcolor{lighterred} 29.2 \\
        DreamO & 71.7 & 70.0 & 70.0 & 70.3 & \cellcolor{lightred} 31.8 & \cellcolor{lightred} 20.7 & \cellcolor{lightred} 27.0 & 45.5 & 52.1 & 40.0 & 51.2 & 43.7 \\
        HiDream-Default & \cellcolor{lightred} 55.7 & \cellcolor{lightred} 60.0 & \cellcolor{lightred} 56.0 & \cellcolor{lightred} 60.0 & 35.7 & \cellcolor{lighterred}  32.0 & 29.3 & \cellcolor{lightred} 32.5 & \cellcolor{lightred} 32.4 & \cellcolor{lightred} 26.2 & \cellcolor{lightred} 28.3 & \cellcolor{lightred} 15.9 \\
        \hline
        DisCo & \cellcolor{lightgreen} 92.0 & \cellcolor{lightgreen} 86.3 & \cellcolor{lightgreen} 95.7 & \cellcolor{lightgreen} 95.7 & \cellcolor{lightgreen} 98.7 & \cellcolor{lightgreen} 98.3 & \cellcolor{lightgreen} 97.7 & \cellcolor{lightgreen} 99.7 & \cellcolor{lightgreen} 100.0 & \cellcolor{lightgreen} 100.0 & \cellcolor{lightgreen} 98.7 & \cellcolor{lightgreen} 98.5 \\
        \bottomrule
    \end{tabular}
    \label{tab:diversity_splits}
\end{table*}

\subsubsection{Grid Search on Reward Weights}

Table~\ref{tab:weight_ablation} presents results from our systematic exploration of reward weight combinations to understand the sensitivity and optimal balance of our compositional reward function. It is on the Flux-Dev baseline. We apply DisCo finetuning for 300 epochs. The analysis reveals that intra-image diversity ($\alpha$) has the strongest impact on overall performance, with higher weights leading to better Unique Face Accuracy and Global Identity Spread. The group-wise diversity component ($\beta$) shows diminishing returns beyond moderate values, while count accuracy ($\gamma$) requires careful balancing to avoid over-penalization. Quality component ($\zeta$) demonstrates that moderate values suffice for maintaining perceptual quality without sacrificing diversity objectives. We pick the optimal configuration $\alpha=0.5$, $\beta=0.1$, $\gamma=0.3$, $\zeta=0.2$ for our final experiment. Note that the final results in Section~\ref{sec:results}(at 480 epochs) are slightly different, as the results in this Table are all compared at 300 epochs to stay consistent.

\begin{table*}[htbp]
    \caption{Ablation study on reward weight parameters. Results are for DisCo(Flux-Dev). Each row shows the effect of different weight configurations on overall performance metrics. Our selected hyperparameter configuration is represented in the \colorbox{lightblue}{Blue} row.}
    \renewcommand{\arraystretch}{1.5}
    \fontsize{8.0pt}{6.75pt}\selectfont
    \centering
    \begin{tabular}{cccc|cccc}
        \hline
        \multicolumn{4}{c|}{Reward Weights} & \multicolumn{4}{c}{Metrics} \\
        $\alpha$ & $\beta$ & $\gamma$ & $\zeta$ & Count & Unique Face & Global Identity & HPS \\
        (Intra-Img) & (Grp-wise) & (Count) & (Quality) & Accuracy & Accuracy (UFA) & Spread (GIS) & \\
        \hline
        
        0.3 & 0.1 & 0.2 & 0.4 & 84.2 & 90.1 & 77.7 & \textbf{33.8} \\
        0.3 & 0.1 & 0.4 & 0.2 & 81.2 & 86.3 & 87.7 & 33.0 \\
        \cellcolor{lightblue} 0.5 & \cellcolor{lightblue} 0.1 & \cellcolor{lightblue} 0.2 & \cellcolor{lightblue} 0.2 & \cellcolor{lightblue} 88.3 & \cellcolor{lightblue} \textbf{96.7} & \cellcolor{lightblue} 97.4 & \cellcolor{lightblue} 33.6 \\
        0.5 & 0.2 & 0.3 & 0.0 & \textbf{90.0} & 95.3 & \textbf{98.1} & 29.3 \\
        0.5 & 0.0 & 0.3 & 0.2 & 87.8 & 94.5 & 80.1 & 33.7 \\
        
        \bottomrule
    \end{tabular}
    \label{tab:weight_ablation}
\end{table*}

\subsubsection{Intra-Image Diversity Aggregation Function Analysis}

Table~\ref{tab:aggregation_ablation} compares different aggregation strategies for computing the intra-image diversity reward when multiple faces are detected within a single image. We perform this analysis on the harder-to-converge DisCo-Krea setup. The results are on DiversePrompts. The choice of aggregation function impacts both convergence behavior and final performance characteristics.

\begin{table}[htbp]
    \caption{Comparison of aggregation functions for intra-image diversity reward computation. Results show performance on Flux-Krea baseline. \colorbox{lightblue}{Blue} represents the selected aggregation function.}
    \renewcommand{\arraystretch}{1.5}
    \fontsize{8.0pt}{6.75pt}\selectfont
    \centering
    \begin{tabular}{l|cccc}
        \hline
        Aggregation & Count & Unique Face & Global Identity & HPS \\
        Function & Accuracy & Accuracy (UFA) & Spread (GIS) & Score \\
        \hline
        
        \cellcolor{lightblue} max() & \cellcolor{lightblue} 83.8 & \cellcolor{lightblue} \textbf{80.1} & \cellcolor{lightblue} \textbf{84.1} & \cellcolor{lightblue} \textbf{32.9} \\
        mean() & \textbf{84.3} & 77.8 & 82.3 & 32.8 \\
        min() & 84.1 & 74.2 & 77.7 & 32.9 \\
        
        \bottomrule
    \end{tabular}
    \label{tab:aggregation_ablation}
\end{table}

Using max() aggregation drives the network toward eliminating the most similar face pair within each image, penalizing any identity overlaps. This approach, particularly when combined with curriculum learning, enables faster convergence and achieves lesser overlapping identities. It essentially implements a "fix the worst violation" strategy that systematically eliminates duplicate identities.

In contrast, mean() aggregation optimizes for low average similarity across all face pairs, which can result in suboptimal solutions where multiple moderate violations persist rather than being eliminated entirely. It converges more slowly and allows identity overlaps to remain, as the model can satisfy the average similarity constraint without addressing individual duplicate pairs. The min() function shows the poorest performance, as it focuses on the least similar pair and provides insufficient pressure to address problematic duplicates. 

\begin{figure*}[h]
  \centering
  \includegraphics[width=0.7\linewidth]{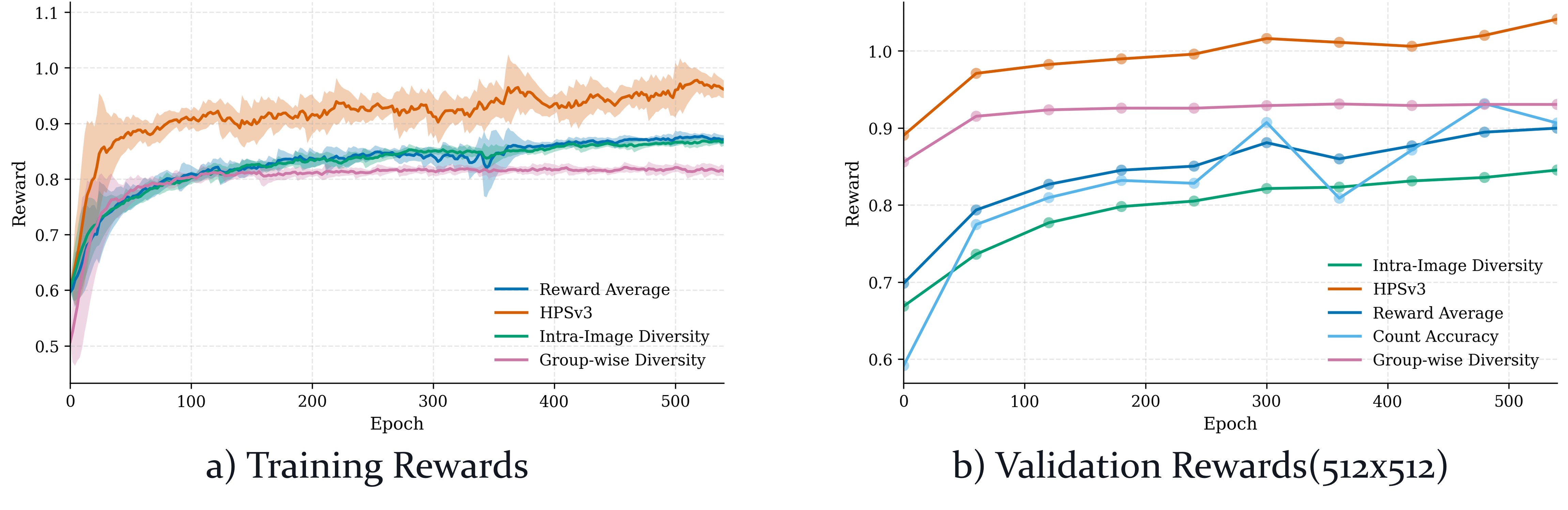}
  \caption{\textbf{DisCo training and evaluation reward curves.} 
As observed, we notice a steady improvement in all four rewards during training and inference.}
  \label{fig:reward_curves}
\end{figure*}

\subsubsection{Reinforcement Learning vs. Supervised Fine-Tuning}
\label{sec:app_sft_vs_rl}

To isolate whether \textsc{DisCo}'s gains stem from RL itself rather than simply additional domain-specific training, we compare against an online SFT baseline: for each group of $M$ generated samples, the highest-reward sample is selected as a supervised target, under the same curriculum schedule and prompt dataset.

\begin{table}[htbp]
    \centering
    \renewcommand{\arraystretch}{1.5}
    \fontsize{8.0pt}{6.75pt}\selectfont
    \setlength{\tabcolsep}{5pt}
    \caption{RL vs. online SFT on Flux-Dev. \textsc{DisCo} outperforms the SFT baseline on identity diversity and perceptual quality, while converging significantly faster.}
    \begin{tabular}{l|ccc}
        \hline
        Model & Unique Face Accuracy (UFA) & HPS Score & Training Time (hrs) \\
        \hline
        Online SFT       & 73.0 & 31.0 & 33 \\
        \rowcolor{lightblue} \textsc{DisCo} (RL) & \textbf{98.6} & \textbf{33.4} & \textbf{13} \\
        \bottomrule
    \end{tabular}
    \label{tab:sft_vs_rl}
\end{table}

As shown in Table~\ref{tab:sft_vs_rl}, \textsc{DisCo} substantially outperforms online SFT on UFA (98.6\% vs.\ 73.0\%) and HPSv2 (33.4 vs.\ 31.0), while converging more than $2\times$ faster (13 vs.\ 33 hours). This gap arises from two factors: RL's exploration and group-normalized advantage estimation encourages diverse generation strategies rather than collapsing to the mode of high-reward samples, and the group-wise diversity reward (Eq.~\ref{eq:group}) exploits GRPO's group structure to penalize cross-sample identity repetition. This is a signal that online SFT cannot capture by optimizing each sample independently.

\subsubsection{Curriculum Bound Ablation}
\label{sec:app_curriculum_ablation}

The curriculum learning strategy anneals from a simple prompt distribution to uniform sampling over the full range. A key design choice is the upper bound of the ``simple'' set. This is the maximum person count used during the initial curriculum phase. Table~\ref{tab:curriculum_ablation} ablates this bound on the Flux-Krea baseline, keeping all other hyperparameters fixed. The no-curriculum row (2-7) corresponds to uniform sampling from the start, equivalent to the penultimate row of Table~2.

\begin{table}[htbp]
    \centering
    \renewcommand{\arraystretch}{1.5}
    \fontsize{8.0pt}{6.75pt}\selectfont
    \setlength{\tabcolsep}{5pt}
    \caption{Ablation on curriculum simple-set bounds on Flux-Krea baseline. \colorbox{lightblue}{Blue} row is the selected configuration.}
    \begin{tabular}{l|cccc}
        \hline
        Curriculum Bound & Count Accuracy & Unique Face Accuracy (UFA) & Global Identity Spread (GIS) & HPS Score \\
        \hline
        2-7 & 79.2 & 82.6 & 73.7 & \textbf{32.4} \\
        2-3 & 84.3 & 83.0 & 85.5 & 32.3 \\
        \rowcolor{lightblue}
        2-4 & \textbf{83.5} & \textbf{89.7} & \textbf{90.6} & 32.2 \\
        2-6 & 80.1 & 83.1 & 77.0 & 32.2 \\
        \bottomrule
    \end{tabular}
    \label{tab:curriculum_ablation}
\end{table}

As shown in Table~\ref{tab:curriculum_ablation}, no curriculum (2-7 uniform) leads to the weakest UFA and GIS, confirming that gradual complexity introduction is critical for the specialist Flux-Krea model. Starting with too narrow a range (2-3) yields strong count accuracy but leaves UFA and GIS below our chosen bound, suggesting insufficient exposure to harder scenarios during the curriculum phase. Starting too broad (2-6) performs similarly to no curriculum on GIS, as the model is exposed to complex prompts too early. The 2-4 bound has the best balance.

\subsubsection{Final Run Reward Curves}

Figure~\ref{fig:reward_curves} demonstrates the training progression of \textsc{DisCo} across all four reward components throughout the learning process. The curves show consistent improvement in intra-image diversity, group-wise diversity, count accuracy, and HPS quality metrics during both training and evaluation phases. While training rewards continue to grow post 500 epochs, the model generates diminishing returns on the testset post 480 epochs. The total training time for a single run is 13 hours.

\subsubsection{Computational Analysis}
Table~\ref{tab:efficiency_analysis} presents a comprehensive comparison of computational efficiency across all evaluated models. We report average performance scores from our multi-human generation benchmarks alongside timing measurements to assess the quality-efficiency trade-off. For proprietary models, we report API response times including network latency, while for open-source models we measure local inference runtime on standardized hardware (NVIDIA H100) for generating a 1024×1024 image with default sampling steps.

\textsc{DisCo} demonstrates an excellent balance between generation quality and computational efficiency. While proprietary models like GPT-Image-1 achieve competitive scores, they incur ongoing API costs and lack deployment flexibility. Gemini-Nanobanana offers faster API responses but with significantly lower generation quality. Among open-source alternatives, \textsc{DisCo} variants significantly outperform existing methods in generation quality while maintaining identical inference times to their respective base models. This makes \textsc{DisCo} particularly attractive for applications requiring both high-quality multi-human generation and practical deployment constraints, offering superior performance without sacrificing efficiency.

\begin{table}[htbp]
    \caption{Computational efficiency comparison across all evaluated models. Average scores are from DiverseHumans-TestPrompts benchmark. Runtimes are measured on NVIDIA H100 for open-source models.}
    \renewcommand{\arraystretch}{1.5}
    \fontsize{8.0pt}{6.75pt}\selectfont
    \centering
    \begin{tabular}{ll|cc}
        \hline
         \multirow{2}{*}{} & \multirow{2}{*}{Model} & Average & API Time \\
        & & Score & (seconds) \\
        \hline
        
        \cellcolor{lightpurple} & \cellcolor{lightpurple} Gemini-Nanobanana & \cellcolor{lightpurple} 60.0 & \cellcolor{lightpurple} 7 \\     
        \multirow{-2}{*}{\cellcolor{lightpurple} Proprietary} & \cellcolor{lightpurple} GPT-Image-1 & \cellcolor{lightpurple} 78.7 & \cellcolor{lightpurple} 28 \\
        
        \hline
         & & Average & Runtime \\
        & & Score & (seconds) \\
        \hline
        & HiDream     & 46.2 & 22 \\
        & Qwen-Image  & 60.1 & 23 \\
        & OmniGen2    & 48.8 & 14 \\
        & Flux        & 56.0 & 9 \\
        \multirow{-5}{*}{Open-Source} & Flux-Krea    & 57.8 & 6 \\
        
        \hline
        & \textsc{DisCo}(Flux)        & 81.7 & 9 \\
        \multirow{-2}{*}{Ours} & \textsc{DisCo}(Krea) & 76.8 & 6 \\
           
        \bottomrule
    \end{tabular}
    \label{tab:efficiency_analysis}
\end{table}

\subsubsection{User Preference Study}
\label{sec:app_user_study}

Our proposed metrics based on face detection and embedding similarity are efficient and reproducible, but it is important to validate that they align with human perception. We conduct a user preference study to assess whether these metrics meaningfully reflect how humans judge identity diversity in generated images.

\paragraph{Setup.} We collect 22 pairs of multi-human images generated by \textsc{DisCo} and baseline Flux, spanning a range of person counts and prompt types. 25 participants were shown each pair and asked to select the image containing less visually similar faces, i.e., the one exhibiting more identity diversity. Participants were given no information about which model produced each image.

\paragraph{Results.}

\begin{figure}[h]
    \centering
    \includegraphics[width=0.5\linewidth]{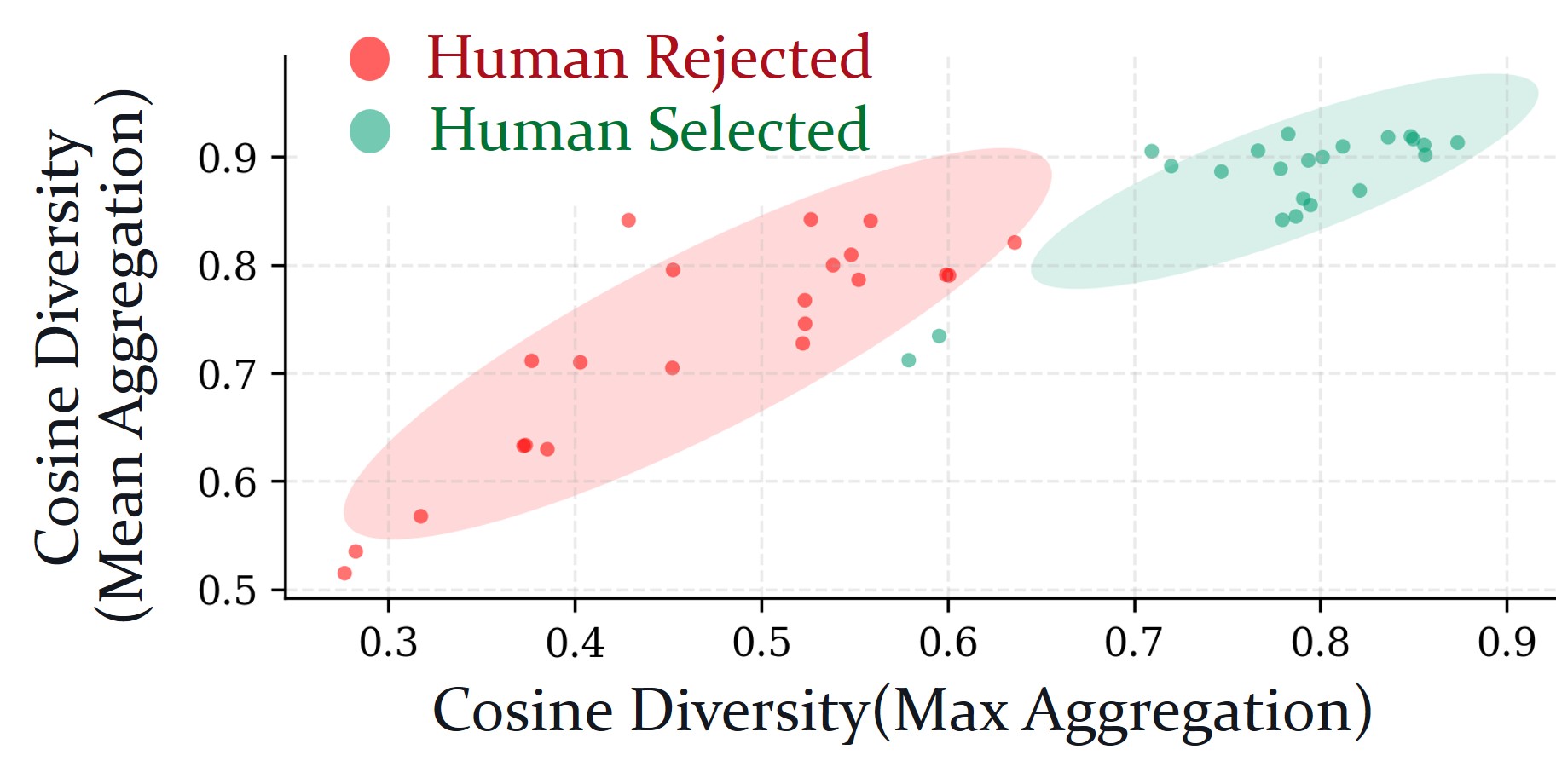}
    \caption{\textbf{User preference study results.} Each point represents one image from a pair, plotted by its mean and max pairwise cosine face similarity. Green points are the images humans collectively judged as more diverse (lower similarity); red points are the images they judged as less diverse. The two clusters separate naturally in the 2D metric space, confirming that mean and max cosine similarity jointly align with human perception of identity diversity.}
    \label{fig:user_study}
\end{figure}

As shown in Figure~\ref{fig:user_study}, the images humans chose as more diverse (green) and less diverse (red) form two naturally separated clusters in the space of mean and max pairwise cosine similarity. Quantitatively, our UFA metric achieves a correlation of \textbf{1.0} with human judgments, and mean pairwise cosine similarity achieves a correlation of \textbf{0.91}, confirming that both the automated metrics and rewards capture what humans actually perceive as identity diversity rather than purely reflecting the behavior of the underlying face recognition model.

\subsection{Qualitative Results}
\subsubsection{Visualizing Global Identity Spread}

Figure~\ref{fig:qualitative_krea} demonstrates the effectiveness of \textsc{DisCo} in achieving global identity diversity compared to the baseline Flux-Dev model. The visualization shows three different prompts, each generating six images using consistent random seeds. The baseline Flux model exhibits significant identity overlap both within individual images and across the generated set, with many faces appearing similar or identical. In contrast, \textsc{DisCo} fine-tuning successfully pushes facial identities apart in the embedding space, resulting in visually distinct individuals across all generations while maintaining high visual quality and prompt adherence.

\begin{figure*}[h]
    \centering
    \includegraphics[width=\textwidth]{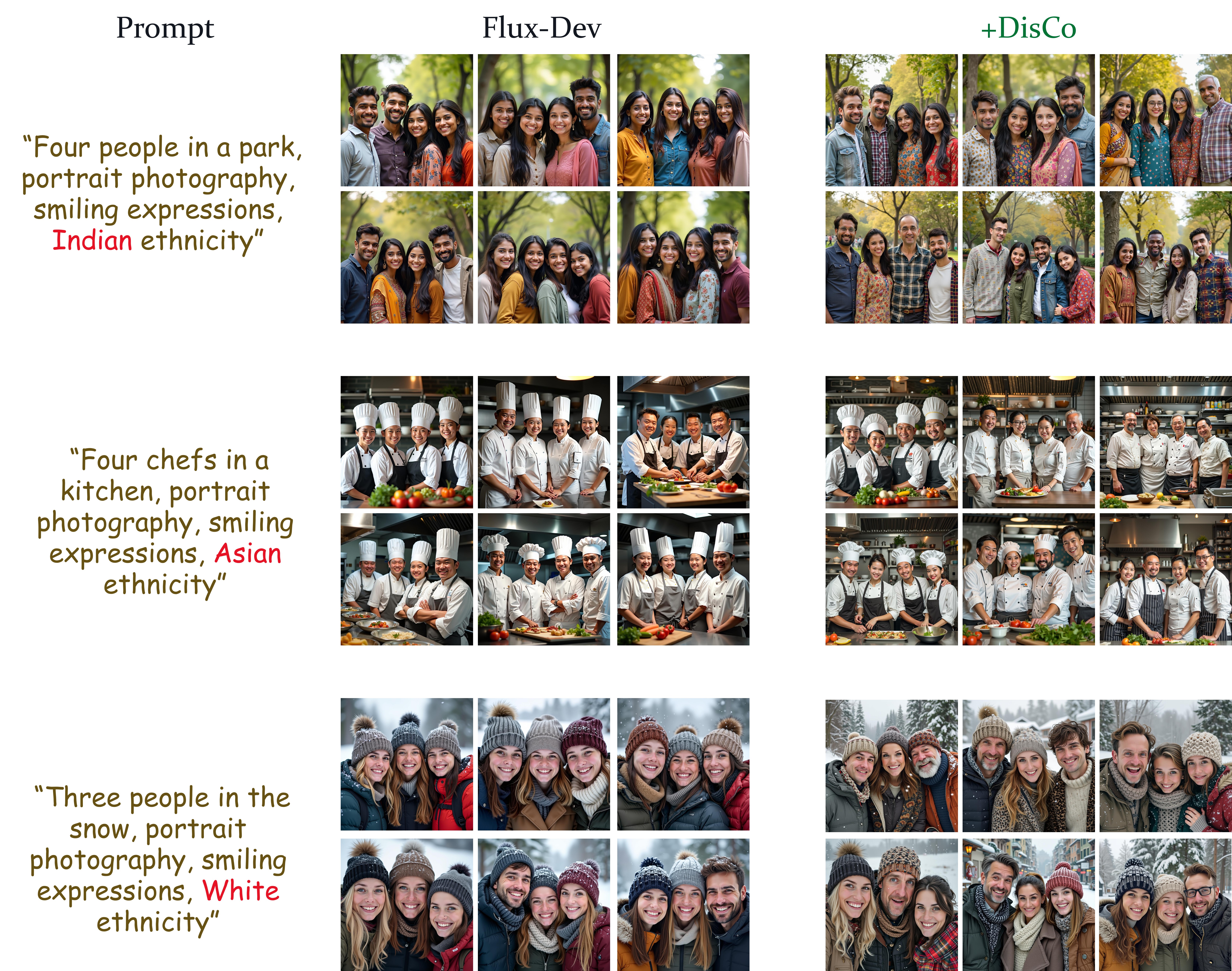}
    \caption{\textbf{\textsc{DisCo} v/s Flux-Dev} As observed in this Figure, we visualize three prompts of people containing the same ethnicity, over six consistent seeds for DisCo and Flux. As observed, Flux results not only generate overlapping identities in the same image, but generate similar looking people across the dataset. However, DisCo finetuning pushes the faces further from each other.} 
    \label{fig:qualitative_krea}
\end{figure*}

\subsubsection{Results on Flux-Krea}

Figure~\ref{fig:qualitative_krea} showcases the qualitative improvements achieved by applying \textsc{DisCo} fine-tuning to the Flux-Krea baseline model. The comparison demonstrates that our approach successfully addresses identity consistency issues present in existing methods while preserving the aesthetic qualities that make Flux-Krea distinctive. The generated images show clear improvements in generating distinct individuals without duplicate identities, accurate person counts matching prompt specifications, and maintained perceptual quality. These results validate that our method generalizes effectively across different base models while preserving their unique characteristics.

\begin{figure*}[h]
    \centering
    \includegraphics[width=\textwidth]{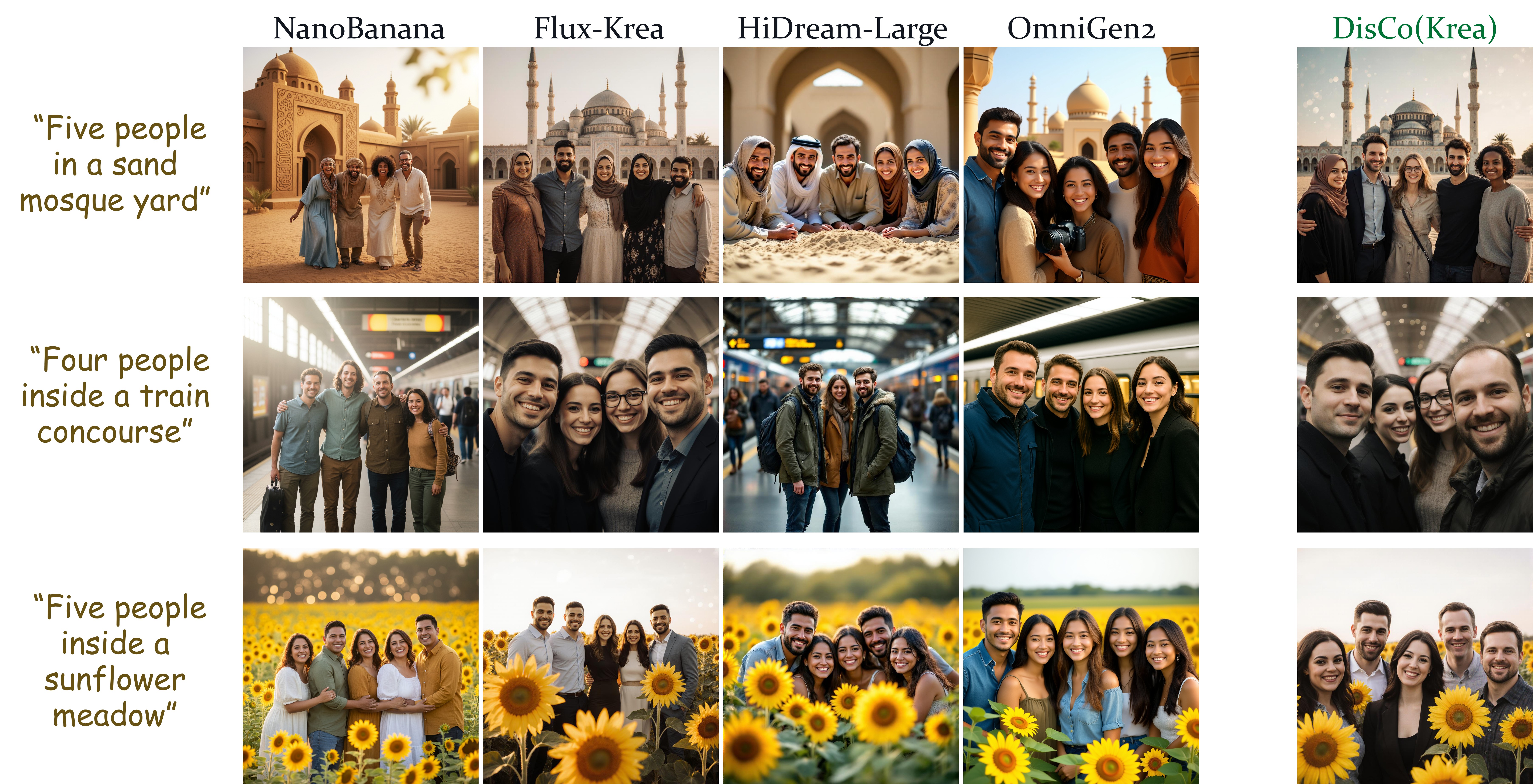}
    \caption{\textbf{\textsc{DisCo-Krea} v/s Related Work} \textsc{DisCo} finetuning applied to Flux-Krea improves performance over current baselines to generate results which consistently generate accurate people without overlapping identity, without a hit in perceptual quality.}
    \label{fig:qualitative_krea}
\end{figure*}

\subsubsection{Visual Effects of Count and HPS Reward Components}

\begin{figure*}[htbp]
    \centering
    \includegraphics[width=0.8\textwidth]{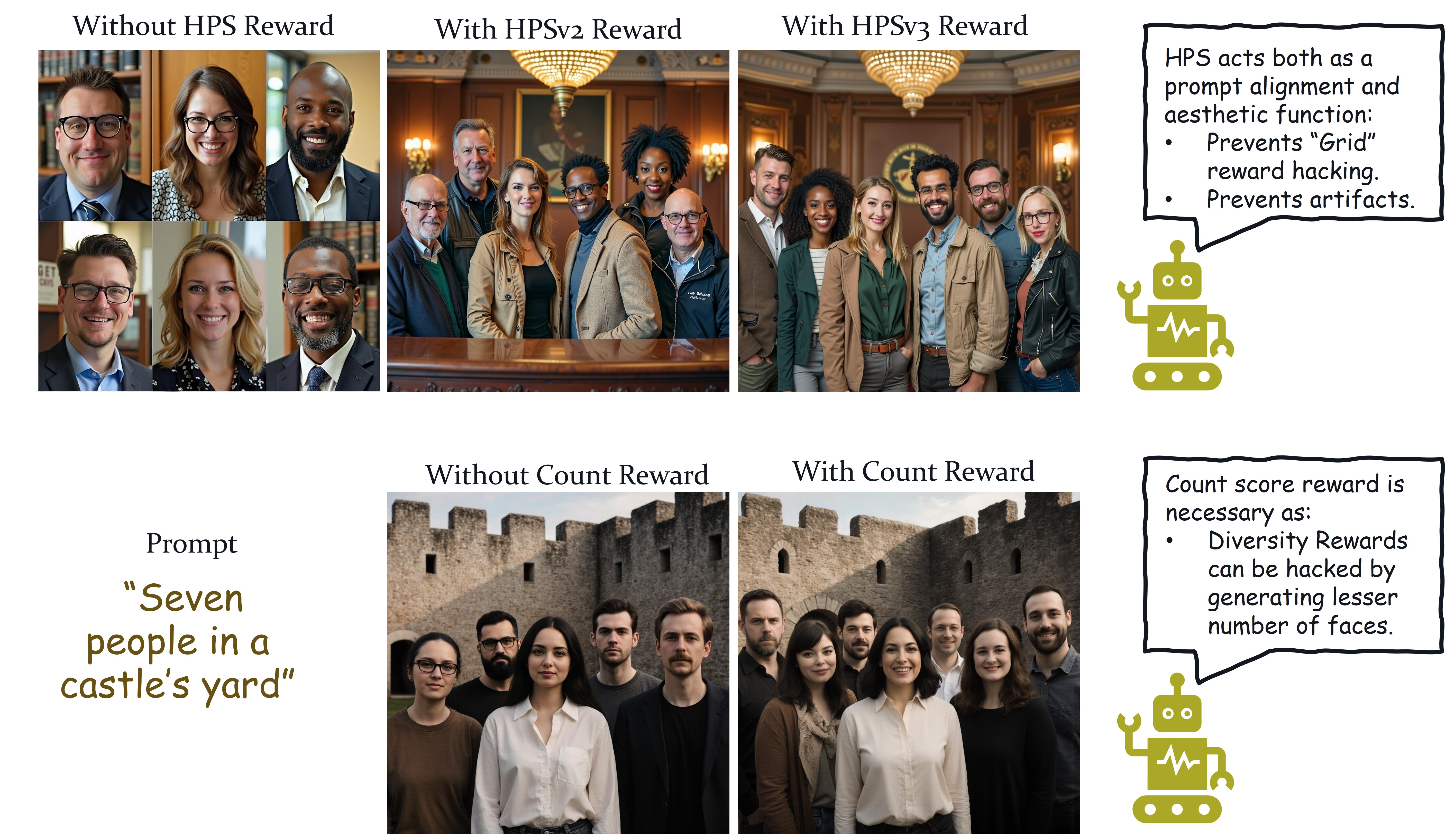}
    \caption{\textbf{Visual effects of count and HPS reward components.} \textit{Top row:} HPS rewards reduce grid artifacts and improve visual quality, with HPSv3 achieving the most natural arrangements. \textit{Bottom row:} Count rewards ensure correct number generation (7 people instead of 5) while maintaining visual coherence.}
    \label{fig:reward_hacking}
\end{figure*}

Figure~\ref{fig:reward_hacking} illustrates the visual effects of our count and HPS reward components in addressing common failure modes during \textsc{DisCo} training. These components are essential for preventing visual artifacts and ensuring realistic multi-person generation.

The top row of Figure~\ref{fig:reward_hacking} demonstrates the visual improvements achieved through HPS rewards. Without perceptual oversight, models produce unnatural grid-like face arrangements that technically satisfy count and diversity requirements but result in unrealistic images. The progression from no HPS to HPSv2 to HPSv3 shows systematic improvement in visual coherence, with HPSv3 producing the most aesthetically pleasing results and minimal degradation artifacts.

The bottom row illustrates the visual impact of count rewards: as shown in Figure~\ref{fig:reward_hacking}, without count control the model generates fewer people than requested (5 instead of 7) to avoid the challenging task of creating multiple distinct identities. Our count reward component directly addresses this by ensuring the correct number of people are generated while maintaining visual quality.

Together, these reward components ensure that our approach produces visually coherent and accurate multi-person generations, preventing both under-generation and visual artifacts that can emerge from optimizing individual objectives in isolation.

\section{Generalizability and Applicability of \textsc{DisCo}}
\label{sec:app_generalizability}

\subsection{Out-of-Distribution Evaluation for number of People}
\label{sec:app_extended_faces}

Our main experiments evaluate on prompts with 2-7 people, matching the training distribution. Here we examine whether \textsc{DisCo} generalizes to out-of-distribution person counts, evaluating on prompts with up to 10 individuals across 270 samples.

\begin{figure}[h]
    \centering
    \includegraphics[width=0.9\linewidth]{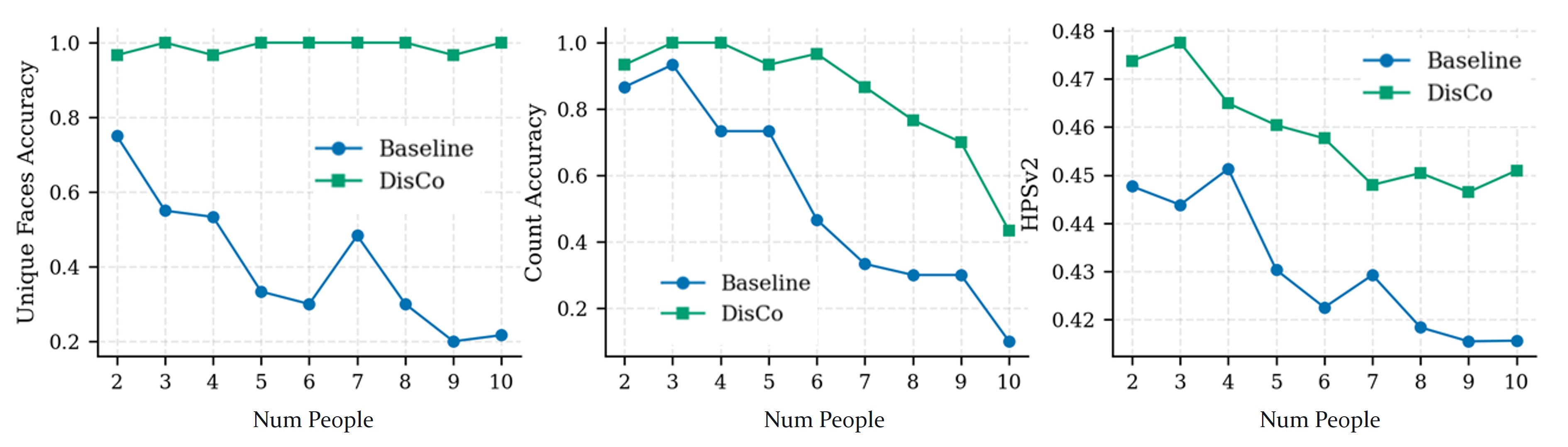}
    \caption{\textbf{Performance on 2-10 people.} \textsc{DisCo}(Flux) retains high Unique Face Accuracy, Count Accuracy and HPSv2 beyond the training range of 7 people, continuing to outperform baseline Flux at 8, 9, and 10 individuals.}
    \label{fig:extended_faces}
\end{figure}

Figure~\ref{fig:extended_faces} shows that \textsc{DisCo} generalizes gracefully beyond its training range. Unique Face Accuracy remains high at 8-10 people, where baseline Flux degrades sharply. This is due to the fact that the identity collapse problem becomes significantly worse as group size grows, yet \textsc{DisCo} continues to produce distinct individuals. Count Accuracy and HPSv2 scores do tend to drop-off for both methods. Regardless, \textsc{DisCo} remains to significantly outperform the baseline at higher person counts. These observations demonstrate that the gains observed within the training range are not simply a product of distribution memorization but reflect a genuine improvement in the model's ability to handle complex multi-human scenes.

\subsection{Beyond Facial Diversity}
\label{app:clothbtypediv}

\textsc{DisCo} is trained with facial identity diversity rewards. In this section, we study whether the learned behavior generalizes to other visual attributes that distinguish individuals in a scene. We investigate two such attributes: clothing diversity and body-type diversity.

\paragraph{Metrics.} \textbf{Clothing diversity} is measured as the average pairwise CLIP~\citep{radford2021learning} similarity between clothing regions across people within each generated image. Clothing regions are obtained by cropping the torso area of each detected person. Lower similarity indicates more diverse clothing across individuals. \textbf{Body-type diversity} is measured as the ratio of people sharing the same somatotype within an image, assessed using Qwen3-VL~\citep{bai2025qwen3} to classify each person's body type from the generated image. A lower ratio indicates greater variety in body types across the generated group.

For both metrics, lower scores indicate better diversity, denoted by $\downarrow$ in Table~\ref{tab:beyond_face}.

\paragraph{Quantitative Results.}

\begin{table}[htbp]
    \centering
    \renewcommand{\arraystretch}{1.5}
    \fontsize{8.0pt}{6.75pt}\selectfont
    \setlength{\tabcolsep}{5pt}
    \caption{Clothing and body-type diversity. \textsc{DisCo} trained on facial diversity already improves both attributes; fine-tuning with a clothing-specific reward (Fashion-SigLIP) yields further gains.}
    \begin{tabular}{l|cc}
    \toprule
    Model & Clothing Similarity$\downarrow$ & Bodytype Similarity$\downarrow$ \\
    \hline
    Flux-Dev                        & 0.73 & 0.55 \\
    \rowcolor{lightblue} +\textsc{DisCo}(Face)  & 0.65 & \textbf{0.49} \\
    \rowcolor{lightblue} +\textsc{DisCo}(Cloth) & \textbf{0.60} & \textbf{0.49} \\
    \bottomrule
    \end{tabular}
    \label{tab:beyond_face}
\end{table}

Table~\ref{tab:beyond_face} shows two findings. First, \textsc{DisCo} trained solely on facial diversity already improves both clothing and body-type diversity over the Flux-Dev baseline. Note that this is despite never explicitly optimizing for either attribute. We hypothesize this is because the model learns a broader notion of identity distinctiveness: generating people whose faces look different from each other naturally induces variation in clothing and body type along with the facial features. Second, replacing the ArcFace identity reward with a Fashion-SigLIP\footnote{https://huggingface.co/Marqo/marqo-fashionSigLIP} clothing similarity signal yields further gains in clothing diversity while retaining the body-type improvement, demonstrating that \textsc{DisCo}'s reward framework generalizes cleanly to other embedding spaces and attribute types beyond faces.

\paragraph{Qualitative results.}
 
\begin{figure}[h]
    \centering
    \includegraphics[width=\linewidth]{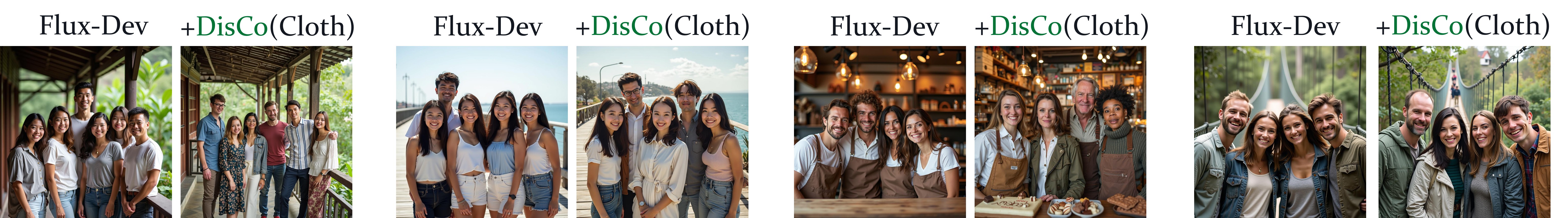}
    \caption{\textbf{Generalization to clothing diversity.} Qualitative comparison between Flux-Dev and \textsc{DisCo}(Cloth). Fine-tuning with a Fashion-SigLIP clothing reward produces noticeably distinct clothes across all generated people.}
    \label{fig:cloth_diversity}
\end{figure}
 
Figure~\ref{fig:cloth_diversity} illustrates these trends qualitatively. The baseline Flux model tends to dress generated individuals in similar or near-identical outfits. \textsc{DisCo} visibly increases clothing variation as a byproduct of identity diversity training, producing groups where each person's clothing is clearly distinct in style, color, and type.

\subsection{General Text-to-Image Performance on GenEval2}
\label{sec:app_geneval}

A valid concern with any RL fine-tuning approach is whether optimizing for a specific objective degrades the base model's general capabilities. We evaluate \textsc{DisCo} on GenEval2~\citep{kamath2025geneval}, a comprehensive benchmark that tests compositional text-to-image generation across object presence, attributes, counting, spatial positioning, and verb understanding, as well as overall prompt complexity (atomicity 3-7+).

\begin{table}[htbp]
    \centering
    \renewcommand{\arraystretch}{1.5}
    \fontsize{8.0pt}{6.75pt}\selectfont
    \setlength{\tabcolsep}{3.5pt}
    \caption{GenEval2 results. \textsc{DisCo} improves over baseline Flux across most compositional attributes and prompt complexity levels, demonstrating that diversity-focused RL fine-tuning does not hurt general text-to-image capability.}
    \begin{tabular}{l|ccccc|cccccc}
    \toprule
    & \multicolumn{5}{c|}{Attribute Type} & \multicolumn{6}{c}{Atomicity} \\
    & Object & Attribute & Count & Position & Verb & 3 & 4 & 5 & 6 & 7+ & \textbf{Mean} \\
    \hline
    Flux-Dev & 89.2 & \textbf{67.2} & 53.9 & 41.5 & 19.0 & \textbf{61.4} & 39.6 & 27.5 & 29.2 & 8.7 & 24.6 \\
    \rowcolor{lightblue} +\textsc{DisCo} & \textbf{94.5} & 66.9 & \textbf{58.7} & \textbf{45.1} & \textbf{27.2} & 61.2 & \textbf{49.2} & \textbf{32.2} & \textbf{29.9} & \textbf{11.0} & \textbf{27.6} \\
    \bottomrule
    \end{tabular}
    \label{tab:geneval}
\end{table}

As shown in Table~\ref{tab:geneval}, \textsc{DisCo} improves over baseline Flux on nearly all axes, including object, counting, positioning, and verb. The mean score rises from 24.6 to 27.6. Gains are particularly notable at higher prompt complexity levels (atomicity 4-7+), which correspond to the multi-human scenarios \textsc{DisCo} is trained on. The only marginal regression is on attribute following (67.2 to 66.9), which we consider negligible. These results indicate that \textsc{DisCo}'s RL fine-tuning using count and HPS rewards not only preserves but actively improves the base model's general compositional capabilities. This is consistent with RL's known ability to correct ingrained model behaviors while retaining broad generalization.

\subsection{Compositional Control with \textsc{DisCo}}
\label{sec:app_composition}

\begin{figure}[h]
    \centering
    \includegraphics[width=0.6\linewidth]{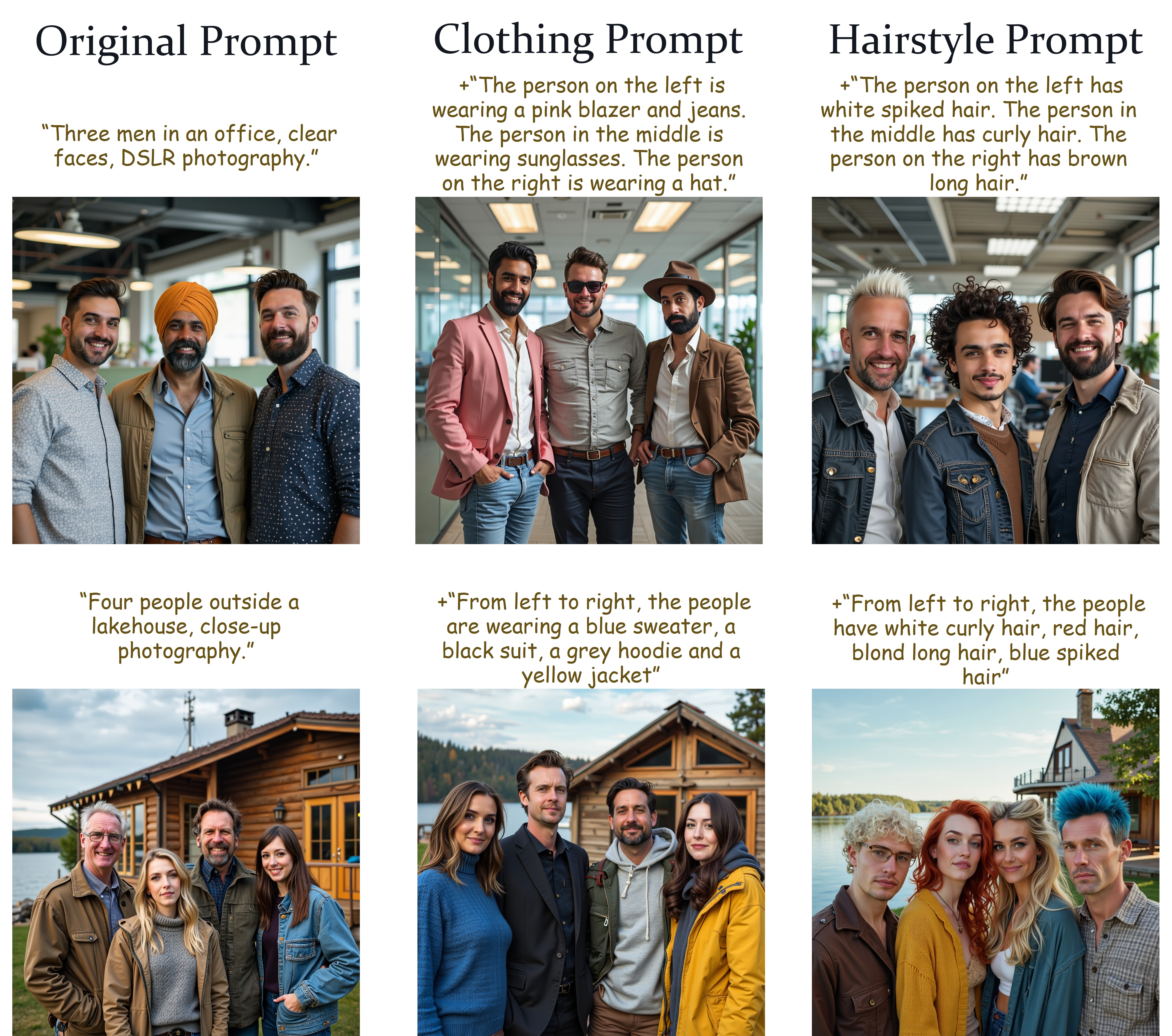}
    \caption{\textbf{\textsc{DisCo} follows per-person compositional prompts.} Despite being trained purely for identity diversity, \textsc{DisCo} successfully generates distinct individuals with specified per-person clothing and hairstyles.}
    \label{fig:composition}
\end{figure}

\textsc{DisCo} is trained for identity diversity, yet we observe that fine-grained compositional control is preserved and in many cases improved. This is due to the HPSv3 reward component. Figure~\ref{fig:composition} illustrates this: prompts with explicit per-person clothing and hairstyle specifications are faithfully followed across all individuals simultaneously.

\paragraph{Quantitative Results.}
Table~\ref{tab:multihuman_generation} corroborates this qualitatively. On both DiverseHumans-Test and MultiHuman-TestBench, which explicitly test complex compositional scenes with per-person action specifications, \textsc{DisCo} retains and slightly improves Action scores over the base model, while also improving HPSv2. This is the harder test of compositional following, as prompts describe distinct actions per individual rather than a shared scene context. We attribute this to the HPSv3 reward, which directly optimizes prompt alignment during RL fine-tuning and prevents quality regression which might otherwise arise from diversity-focused optimization. This is further supported by our GenEval2 results (Table~\ref{tab:geneval}), where \textsc{DisCo} improves over baseline Flux on compositional attributes including count, position, and verb understanding. Future work such as Ar2Can~\citep{borse2025ar2can} builds on this foundation toward more explicit per-person spatial and identity-level control.

\section{Limitations and Future Work}
\label{sec:limitations}
\textsc{DisCo} relies on face detection and embedding similarity, so performance can degrade under heavy occlusion, extreme poses, or very small faces. While \textsc{DisCo} does not explicitly train for per-person compositional attributes such as clothing or body type, we show in Appendix~\ref{sec:app_composition} that the HPSv3 reward preserves and often improves the base model's ability to follow such specifications. DisCo is not a personalization method and only text-to-image generation. However, personalization using explicit spatial and identity-level control is explored in concurrent work Ar2Can~\citep{borse2025ar2can}, which builds directly on \textsc{DisCo}'s RL formulation. Other future directions include extending diversity optimization to whole-body appearance cues, video generation with spatiotemporal identity consistency, and more explicit fairness and demographic balance objectives.

\section{Ethics Statement}
\label{sec:ethics}

Our work focuses on improving identity diversity in multi-human text-to-image generation to enhance fairness and realism in generative models. No human subjects, images or real identities were used; all experiments relied on (sanitized) text prompts and synthetic data. We anticipate positive societal benefits from our advancements in AI-driven multi-human image generation. By developing models that accurately generate diverse individuals across age, ethnicity, and gender, we aim to contribute to more equitable and inclusive digital media. Our work can enhance creative tools for artists and developers, enrich AR/VR/XR experiences, and improve assistive technologies. At the same time, we recognize potential risks, including misuse for misinformation campaigns or for impersonation. We also disclose the use of large language models (LLMs) for prompt generation, formatting assistance(for tables, plots), and text refinement. All generated outputs were carefully reviewed for quality and accuracy, and the scientific contributions, experiments, and conclusions remain the original work of the authors. We emphasize the importance of transparency, fairness audits, and responsible release practices, and strongly discourage malicious applications of this technology.

\section{Reproducibility}
\label{sec:reproduce}

To ensure reproducibility, we provide comprehensive implementation details as part of this submission. Our DISCO framework is implemented on top of the publicly available Flow-GRPO codebase, with training configurations specified in Appendix~\ref{sec:imp_det} (480 epochs, learning rate $1 \times 10^{-4}$, compositional reward weights $(\alpha=0.50, \beta=0.10, \gamma=0.15, \zeta=0.15)$, and curriculum parameters $(\gamma=2.0, t_{\text{curriculum}}=40{,}000$ steps)). Appendix~\ref{app:ext_method} provides complete algorithmic descriptions and pseudocode for group-wise diversity computation (Algorithm~\ref{alg:group_diversity}), curriculum learning (Algorithm~\ref{alg:curriculum}), and the full DISCO training procedure (Algorithm~\ref{alg:disco}). We also reference the publicly available detector and face embedding models. Our training dataset and the DiverseHumans evaluation set of 1,200 prompts are described in Appendix~\ref{sec:appendix_datasets}, along with the (publicly available) MultiHuman-TestBench dataset used for evaluation. All evaluation metrics (Count Accuracy, Unique Face Accuracy, Global Identity Spread) are mathematically defined in Appendix~\ref{app:metrics}, with explicit similarity thresholds $(\kappa_{\text{dup}}=0.5)$ and clustering procedures. Baseline model evaluations follow official hyperparameters as documented in Appendix~\ref{sec:imp_det}, ensuring fair comparison. Finally, our distributed training setup (21 H100 GPUs with specified batch sizes and gradient accumulation) is fully documented in Appendix~\ref{sec:imp_det} to facilitate replication of our results.

\end{document}